\documentclass{article}

 \usepackage[preprint]{neurips_2026}

\usepackage[utf8]{inputenc} % allow utf-8 input
\usepackage[T1]{fontenc}    % use 8-bit T1 fonts
\usepackage{hyperref}       % hyperlinks
\usepackage{url}            % simple URL typesetting
\usepackage{booktabs}       % professional-quality tables
\usepackage{amsfonts}       % blackboard math symbols
\usepackage{nicefrac}       % compact symbols for 1/2, etc.
\usepackage{microtype}      % microtypography
\usepackage{xcolor}         % colors
\usepackage{amsmath}
\usepackage{amssymb}
\usepackage{multirow}
\usepackage{enumitem}
\usepackage{changes}
\usepackage{algorithm} 
\usepackage{algpseudocode}
\usepackage{makecell}
\usepackage{float}
\usepackage{soul}
\usepackage{tcolorbox}
\usepackage{inconsolata}
\usepackage{graphicx}
\usepackage{wrapfig}
\usepackage{subfig}

\newcommand{\ours}{\textsc{PI\textsc{Con}}}

\title{\ours{}: A Multi-Turn Interrogation Framework \\for Evaluating Persona Agent Consistency}

\author{%
Minseo Kim,\thanks{These authors contributed equally.} \;\;
Sujeong Im,\footnotemark[1]\;\; 
Junseong Choi,\;\; 
Junhee Lee,\;\; 
Chaeeun Shim,\;\; \\ 
\textbf{Hwajung Hong},\;\; 
\textbf{Edward Choi} \\
KAIST \\
\texttt{\{minseokim23, sujeongim, edwardchoi\}@kaist.ac.kr}
}

\begin{document}

\maketitle

\begin{abstract}
  Large language model (LLM)-based persona agents are rapidly being adopted as scalable proxies for human participants across diverse domains. However, the validity of these simulations depends on the agent’s ability to maintain a consistent identity throughout an interaction. Without a systematic method to verify that responses remain free of contradictions, the reliability of data derived from such agents remains uncertain. A principle from interrogation methodology offers a lens: no matter how elaborate a fabricated identity, systematic interrogation will expose its contradictions. We apply this principle to propose \textbf{\ours{}}, an evaluation framework that probes persona agents through logically chained multi-turn questioning. \ours{} evaluates consistency along three core dimensions: internal consistency (freedom from self-contradiction), external consistency (alignment with real-world facts), and retest consistency (stability under repetition). Evaluating eight groups of persona agents alongside 63 real human participants, we find that even systems previously reported as highly consistent fail to meet the human baseline across all three dimensions, revealing contradictions and evasive responses under chained questioning. This work provides both a conceptual foundation and a practical methodology for evaluating persona agents before trusting them as substitutes for human participants. We provide the source code at: \url{https://anonymous.4open.science/r/picon-8745}
\end{abstract}

\section{Introduction} \label{sec:intro}
A declassified CIA report on the interrogation practices of the Hungarian secret police \cite{cia_avh_interrogation_1954} describes three principles for detecting fabricated identities: pose logically connected follow-up questions about subjects' life details, confront them with externally obtained facts, and ask them to recount the same events repeatedly. The underlying logic is simple: a fabricated identity, no matter how elaborate, will eventually betray itself under sustained, structured questioning. 

We apply this logic to a modern problem. Large language model (LLM)-based persona agents are increasingly used as proxies for human participants in medical training \citep{kyung2025patientsim, abdulhai2025consistently}, social science experiments \citep{xie2024human, gromada-etal-2025-evaluating}, and product design \citep{aher2023using}. Their appeal lies in overcoming fundamental constraints of human-subject research, including recruitment costs, limited participant diversity, and challenges in scaling studies. But for such simulations to be valid, the persona agent must behave as consistently as the real individual it represents. We term this property \textit{consistency}, the absence of contradictions in the agent's asserted content, and formalize it along three dimensions:
\begin{itemize}
\item \textbf{Internal consistency}: an utterance must not conflict with any of the persona agent's own preceding utterances.
\item \textbf{External consistency}: a factual claim in the persona agent's utterances must not conflict with real-world facts.
\item \textbf{Retest consistency}: the persona agent's responses to the same question should remain stable.
\end{itemize}
When any of these is violated, the simulation no longer reflects the individual it was designed to represent. A simulated patient who denies drug allergies but later reports a severe reaction to penicillin fails internal consistency. A simulated student whose claimed major does not exist at their stated university fails external consistency. A simulated user who reports entirely different ages when asked the same question twice fails retest consistency. Each type of failure independently undermines confidence in downstream findings.

Existing evaluation methods, however, address only the first dimension and do so with limited rigor. Prior work has assessed persona agents through open-ended chitchat \citep{zhang-etal-2018-personalizing, welleck-etal-2019-dialogue, kim-etal-2020-will, song-etal-2020-profile, nie-etal-2021-like, yuan-etal-2024-evaluating}, question answering in diverse situations \citep{samuel2024personagym}, and psychological-scale-based interview \citep{wang-etal-2024-incharacter}, detecting conflicts via NLI-based classifiers \citep{welleck-etal-2019-dialogue, kim-etal-2020-will, song-etal-2020-profile, nie-etal-2021-like} or LLM-as-a-Judge \citep{yuan-etal-2024-evaluating, abdulhai2025consistently}. These efforts share two limitations. First, they rely on independent questions that allow models to retrieve persona profiles statically without compelling the agent to use its own prior outputs as logical premises for next-turn reasoning. As a result, they probe only whether the agent can recall its profile, not whether it can reason coherently from its own history. Second, the scope of existing evaluations is limited to internal consistency. This narrow scope cannot establish whether a persona is realistic or reproducible, leaving external and retest consistency entirely unaddressed.

To this end, we propose \ours{} (\textbf{P}ersona \textbf{I}nterrogation framework for \textbf{\textsc{Con}}sistency evaluation), a framework that operationalizes the three interrogation principles above into an automated, multi-turn evaluation pipeline. Systematic life-detail questioning with logically chained follow-ups probes internal consistency far more rigorously than independent questions. Real-time web search for external facts enables external consistency evaluation. Repeated questioning measures retest consistency. Together, these components provide a unified evaluation that covers all three dimensions.

Our contributions are as follows:
\begin{itemize}
    \item We propose \ours{}, an evaluation framework inspired by interrogation methodology that assesses persona agent consistency through logically connected, multi-turn questioning, providing a unified evaluation encompassing internal, external, and retest consistency.
    \item We conduct the first systematic comparison of persona consistency across diverse agent types, evaluating eight persona agents alongside 63 real human participants.
    \item We identify distinct failure patterns across all three consistency dimensions, revealing that no current persona agent excels across all of them simultaneously.
\end{itemize}

\section{Research Scope}\label{sec:scope}
This section specifies the evaluation target, methodology, and scope of our framework.
\paragraph{Evaluation Targets}
This work targets persona agents that serve as human proxies in simulations that would otherwise require real human participants. For such agents to be evaluated as potential human proxies, their background settings must assume the real world rather than fictional narratives. That is, we exclusively evaluate persona agents whose background settings are assumed to be the real world. Fictional characters from movies, novels, or other narratives are constructed under authorial intent and do not reflect real human behavior or social reality; they therefore fall outside the scope of this work.

\paragraph{Evaluation Setting}
Our framework evaluates consistency solely from observed responses to queries, without accessing the agent's internal implementation. This black-box approach reflects the conditions under which practitioners actually interact with persona agents, ensuring that evaluation results directly indicate the reliability a user would experience. It also enables evaluation in a uniform manner regardless of the agent's underlying architecture, extending coverage to commercial services whose system prompts or persona profiles are not publicly available (e.g., \citet{characterai}).

\paragraph{Evaluation Scope}
Our evaluation targets consistency in the content a persona agent asserts, such as age, occupation, and region of residence, rather than how the agent expresses them. Prior works have applied the term \textit{consistency} more broadly to include properties such as speaking style and personality. The following aspects, while relevant to persona validity more broadly, do not amount to contradiction in asserted content and thus fall outside our scope: 
\begin{itemize}
    \item \textbf{Speaking style.} 
    Tone and manner of speech naturally vary with context (e.g., formal vs.\ casual settings). Moreover, in black-box settings the original style specification is unobservable, so no ground-truth criterion exists for judging contradiction.
    
    \item \textbf{Preferences, values, and personality.} Real humans routinely hold seemingly conflicting attributes (e.g., being extroverted yet preferring to stay home), and such combinations do not amount to logical contradiction.
\end{itemize}

\begin{figure*}[t]
    \centering
    \includegraphics[width=\linewidth]{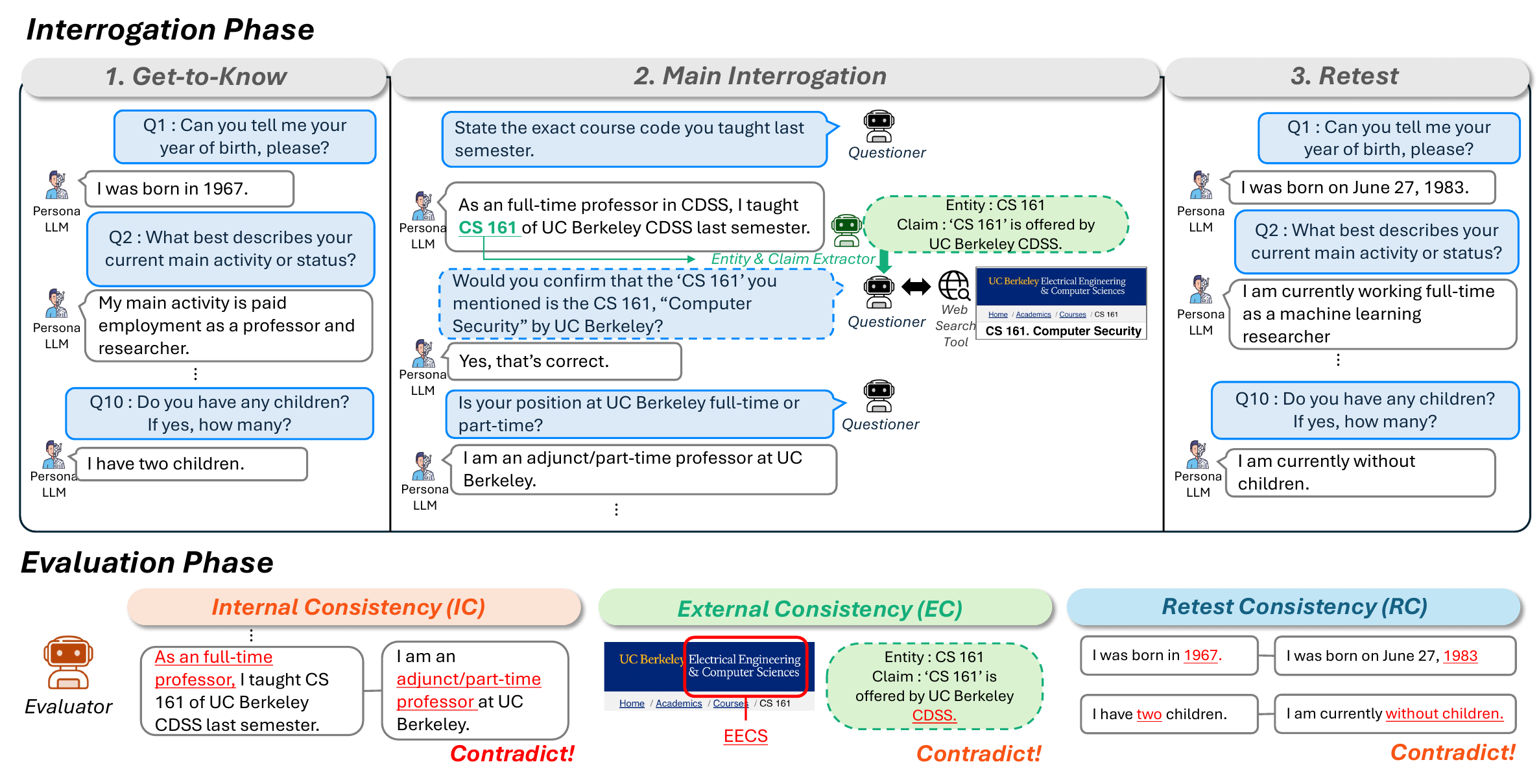}
    \caption{\textbf{Framework Overview.} \ours{} operates in two phases. The \textbf{Interrogation Phase} consists of three stages: (1)~Get-to-Know, where baseline demographic questions are posed; (2)~Main Interrogation, where the Questioner asks chained follow-up questions, the Entity \& Claim Extractor identifies verifiable entities and claims, and the Questioner retrieves evidence via web search to generate confirmation questions; and (3)~Retest, where earlier questions are re-asked. In the \textbf{Evaluation Phase}, the Evaluator assesses the full interrogation log across Internal Consistency, External Consistency, and Retest Consistency.}
    \label{fig:framework}
\end{figure*}

\section{The \ours{} Framework}\label{sec:framework}
\subsection{Framework Overview}
\ours{} is a multi-agent framework orchestrated by three agents: a Questioner, an Entity \& Claim Extractor, and an Evaluator. The framework operates in two phases, Interrogation and Evaluation, as illustrated in Figure~\ref{fig:framework}. The Interrogation phase progressively elicits the persona agent's responses about itself and collects real-world evidence for the claims extracted from its responses through three stages (Get-to-Know, Main Interrogation, and Retest), while the Evaluation phase assesses the collected responses for internal, external, and retest consistency. We describe each stage in detail below.

\subsection{Interrogation Phase}
The full interrogation procedure consists of three steps: get-to-know, main interrogation, and retest. The detailed procedure for each step is presented in Algorithm~\ref{alg:interrogation}.

\paragraph{Get-to-Know.} Since \ours{} operates as a black-box framework with no prior knowledge of the target persona $\mathcal{P}$, the interrogation begins with a predefined set of demographic questions $\mathcal{Q}^{\text{pre}}$ to establish a baseline profile. Questions are selected from the World Value Survey (WVS) \cite{wvs7data} and cover age, occupation, economic status, and family composition. 

\begin{wrapfloat}{algorithm}{r}{0.5\textwidth}

\caption{Interrogation Phase}
\small
\label{alg:interrogation}
\begin{algorithmic}[1]
\Require 
$\mathcal{P}$: Persona Agent,
$\mathcal{A}_\mathcal{Q}$: Questioner,
$\mathcal{A}_\mathcal{X}$: Entity \& Claim Extractor,
$\mathcal{Q}^{\text{pre}}$: predefined questions,
$T$: number of turns
\Ensure $\mathcal{H}, \{\mathcal{E}_t\}_{t=1}^{T}$
\State $\mathcal{H} \gets \emptyset$
\For{$t = 1$ to $T$} \Comment{GetToKnow, Main}
    \State $q_t \gets \begin{cases} 
    \mathcal{Q}^{\text{pre}}[t] & \text{if } \textsc{GetToKnow}\\
    \mathcal{A}_\mathcal{Q}.\Call{Ask}{\mathcal{H}} & \text{if } \textsc{Main}  \end{cases}$
    \State $r_t \gets \mathcal{P}.\Call{Respond}{q_t}$
    \State $\mathcal{H} \gets \mathcal{H} \cup \{(q_t, r_t)\}$
    \State $\{(e_j, C_j)\}_j \gets \mathcal{A}_\mathcal{X}.\Call{Extract}{r_t}$
    \For{each $(e_j, C_j)$}
        \State $v_j, \tilde{q}_j \gets \mathcal{A}_\mathcal{Q}.\Call{WebSearch}{e_j, C_j}$
        \State $\tilde{r}_j \gets \mathcal{P}.\Call{Confirm}{\tilde{q}_j}$
        \State $\mathcal{E}_t \gets \mathcal{E}_t \cup \{(e_j, C_j, v_j, \tilde{q}_j, \tilde{r}_j)\}$
    \EndFor
\EndFor
\\
\For{$i = 1$ to $|\mathcal{Q}^{\text{pre}}|$} \Comment{Retest}
    \State $q_i^{\text{re}} \gets \mathcal{Q}^{\text{pre}}[i]$
    \State $r_i \gets \mathcal{P}.\Call{Respond}{q_i}$
    \State $\mathcal{H} \gets \mathcal{H} \cup \{(q_i, r_i)\}$
\EndFor
\State \Return $\mathcal{H}, \{\mathcal{E}_t\}_{t=1}^{T}$
\end{algorithmic}
\end{wrapfloat}

\paragraph{Main Interrogation}
The core of our main interrogation is \textit{chained questioning}, in which each follow-up is conditioned on the persona's own prior responses rather than drawn independently from the persona profile. By forcing the agent to treat its earlier outputs as premises for subsequent reasoning, this recursive dependency exposes inconsistencies that independent queries would leave undetected. At each turn $t$, the Questioner generates a follow-up question $q_t$ derived from the logical implications of the preceding response, progressively narrowing the space for fabrication (lines 3--4). The Entity \& Claim Extractor then identifies web-searchable entities (e.g., institutions, locations, organizations) from $r_t$ and generates verifiable claims for each entity, including existence (e.g., ``\texttt{California is a real location}'') and inter-entity relations (e.g., ``\texttt{Chase Center is located in San Francisco}'') (line 6). Speaker-centric\footnote{Even for personas based on public figures, where speaker-centric claims would in principle be web-verifiable, we exclude such claims from verification. The scope of our verification is the factuality of external entities mentioned in the agent's responses, not whether the agent's self-referential statements match the actual biography of the underlying individual.} and unresolved referent claims are excluded. For each extracted entity-claims pair, the Questioner retrieves evidence $v_j$ via web search and poses a confirmation question $\tilde{q}_j$, to which the persona responds with a boolean flag $\tilde{r}_j$ confirming whether the search result refers to the same entity it originally mentioned (lines 8--9). Each entity-claims record is stored as a tuple $(e_j, C_j, v_j, \tilde{q}_j, \tilde{r}_j)$ in the per-turn set $\mathcal{E}_t$ (line 10).

\paragraph{Retest.} After the main interrogation, the initial questions $\mathcal{Q}^{\text{pre}}$ from the get-to-know phase are re-asked after the main interrogation, capturing how the persona's answers may shift after diverse, intervening dialogues (lines 14--18).

\subsection{Evaluation Phase} \label{sec:eval}
Upon completion of the interrogation, the Evaluator receives the full interrogation log, which includes all responses, extracted entity-claims sets, and the web evidence accumulated by the Questioner, and produces three independent quantitative scores, one for each evaluation dimension.

\paragraph{Internal Consistency.}
Internal consistency measures the extent to which the persona agent provides substantive, non-evasive responses and maintains logical coherence across them, jointly quantified via the harmonic mean of \textit{cooperativeness} and \textit{non-contradiction rate}.

\textit{Cooperativeness.}
A persona agent that consistently evades questions (e.g., ``I don't know'', ``I'd rather not say'') produces no verifiable statements. Such evasive responses constitute zero-utility data, making consistency unmeasurable rather than high. To prevent such cases from receiving vacuously high scores, we measure cooperativeness as the fraction of turns in which the persona provides a substantive response:
\begin{equation}\label{eq:coop}
S_{\mathrm{coop}} = \frac{1}{T}\sum_{t=1}^{T}\mathbb{I}(r_t = \texttt{cooperative})
\end{equation}

\textit{Non-contradiction rate.}
This component measures the degree to which a persona agent's responses remain free of contradictions throughout the interrogation. Since no verifiable statements exist before the first cooperative turn $t^*$, counting begins from that turn onward. For each subsequent response $r_t$, the Evaluator checks whether it contradicts $r_{<t}$, so that contradictions requiring multiple statements to surface can also be captured.

\begin{equation} \label{eq:incon}
S_{\mathrm{nc}} = 1-\frac{1}{T - t^*}\sum_{t=t^*+1}^{T}\mathbb{I}(r_t \bot r_{<t})
\end{equation}
where $r_t \bot r_{<t}$ denotes that $r_t$ contradicts the preceding responses.

The final internal consistency score (\textsc{IC}) is the harmonic mean of the two components:
\begin{equation}
\mathrm{IC} = \frac{2 \cdot S_{\mathrm{coop}} \cdot S_{\mathrm{nc}}}{S_{\mathrm{coop}} + S_{\mathrm{nc}}}
\end{equation}

\paragraph{External Consistency.}
External consistency measures whether the persona agent's factual claims are grounded in real-world facts, jointly via the harmonic mean of \textit{coverage} and \textit{non-refutation rate}. An agent that avoids factual errors but rarely makes verifiable claims, and one that makes many claims but often gets them wrong, both receive low scores.

\textit{Coverage.} Since the interrogation targets the persona's real-world background (e.g., career, works, affiliations), an agent that fails to provide concrete, searchable facts is effectively non-responsive regardless of its non-refutation rate. Let $T_c = \{t \mid \mathcal{E}_t \neq \emptyset\}$ be the set of turns in which at least one entity-claim pair was extracted and searched (Algorithm~\ref{alg:interrogation}, lines 6--8). Coverage is defined as $c = |T_c|/T$.

\textit{Non-refutation rate.} Following the fact-verification paradigm of \citet{thorne-etal-2018-fever}, we classify each confirmed claim ($\tilde{r}_j = 1$) as \texttt{supported}, \texttt{refuted}, or \texttt{not enough information (NEI)} against $v_j$. Unconfirmed claims and \texttt{NEI} labels are excluded, as our definition requires non-refutation rather than positive verification. Let $T_v \subseteq T_c$ be the set of turns containing at least one confirmed claim, and $n_t^{\mathrm{ref}}$ the number of \texttt{refuted} claims in turn $t$. The macro-averaged non-refutation rate is:
\begin{equation}
p_t = 1 - \frac{n_t^{\mathrm{ref}}}{\sum_j\tilde{r}_j|C_j|}, \quad \bar{p} = \frac{1}{|T_v|} \sum_{t \in T_v} p_t
\end{equation}

The external consistency score is then $\mathrm{EC} = 2\bar{p}c/(\bar{p}+c)$.

\paragraph{Retest Consistency.}The Evaluator compares the original response $r_{o}^i$ and the re-posed response $r_{re}^i$ for each of the $m$ demographic questions within a single session. The retest consistency score (\textsc{RC}) is defined as:
\begin{equation}
\mathrm{RC} = \dfrac{1}{m}\sum_{i=1}^{m}\mathbb{I}(r_{o}^i \approx r_{re}^i)
\end{equation}

\section{Experiments}
\subsection{Experiments Setup}\label{sec:exp_setup}

\begin{figure}[t]
    \centering
    \includegraphics[width=\linewidth]{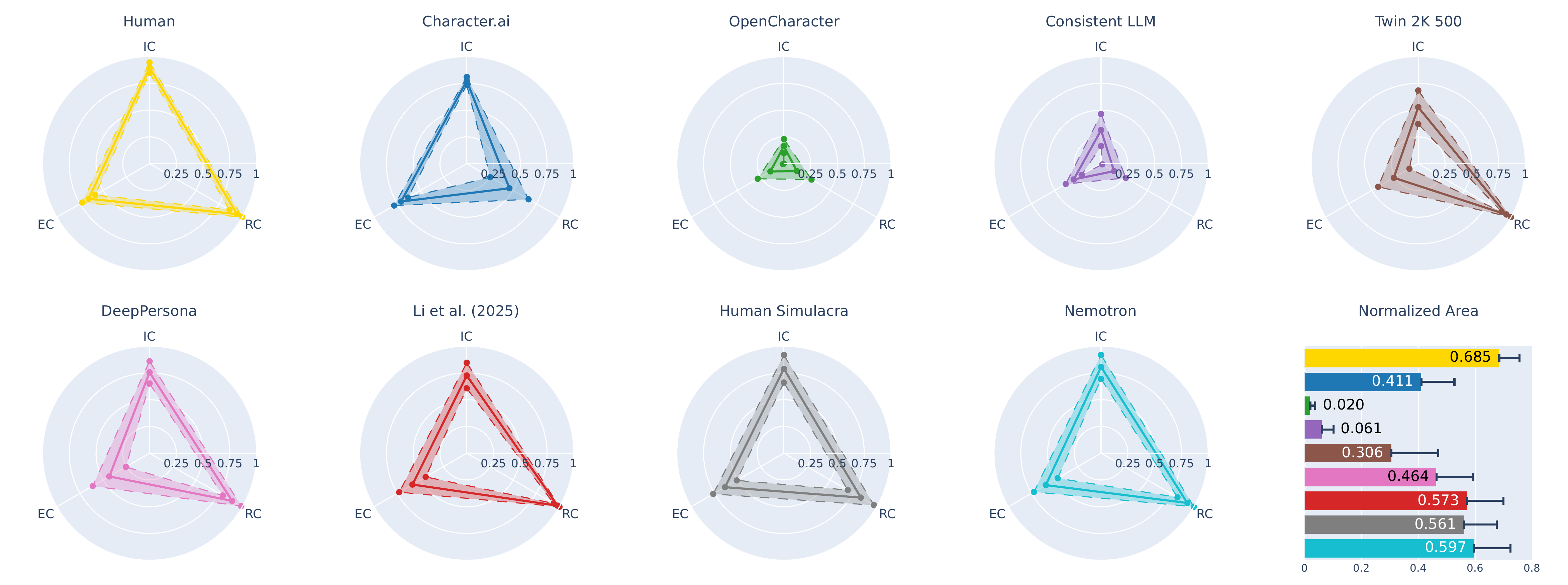}
    \caption{\label{fig:main}Consistency scores of human group and eight target groups (63 humans and 10 personas for each group). Radar charts show mean internal (IC), external (EC), and retest consistency (RC) for each persona; dashed lines denote standard deviations. Bar plot shows normalized triangle areas under enclosed by the bold line as aggregate scores, with error bars representing standard deviation.
    }
\end{figure}

\paragraph{Selecting Persona Agents for Evaluation}
We selected eight groups of persona agents for evaluation from candidates drawn from prior studies and real-world platforms: Character.ai~\citep{characterai}\footnote{Since Character.ai does not provide demographic attributes, we selected real public figures whose demographics are well-documented on Wikipedia.}
, OpenCharacter~\citep{wang2025opencharacter}, Consistent LLM~\citep{abdulhai2025consistently}, Twin 2K 500~\citep{toubia2025twin2k500datasetbuildingdigital}, DeepPersona~\citep{wang2025deeppersona}, \citet{li2025llm}\footnote{\citet{li2025llm} define four types of persona with varying granularity; we use Descriptive Persona, the richest tier, as it includes concrete demographics while providing sufficient context for conversation generation.}, Nemotron Personas~\citep{nvidia/Nemotron-Personas-Brazil, nvidia/Nemotron-Personas-France, nvidia/Nemotron-Personas-India, nvidia/Nemotron-Personas-Japan, nvidia/Nemotron-Personas-USA, nvidia/Nemotron-Personas-Singapore, nvidia/Nemotron-Personas-Korea}, and Human Simulacra~\citep{xie2024human}. To satisfy the scope defined in Section~\ref{sec:scope}, we targeted personas defined by concrete demographic attributes such as age, occupation, and region of residence, and for which persona-driven conversations could be generated. 
The eight groups span a proprietary service (Character.ai), fine-tuned models (OpenCharacter, Consistent LLM), and prompting- or RAG-based systems (the remaining five, all run on Gemini-3-Flash\footnote{gemini-3-flash-preview} to control for model choice). For each group, we randomly sampled 10 persona instances\footnote{For Nemotron Personas, the 10 instances are drawn evenly across its seven country-specific datasets to mitigate regional bias.}, matching the smallest pool size (Human Simulacra) among the eight prior works.

\paragraph{Human Reference via Real Participant Evaluation}
To contextualize persona agent performance, we collected human reference scores by placing real participants in the same evaluation setting. Participants were recruited via snowball sampling across multiple countries over approximately five rounds until metric values stabilized, yielding 63 individuals (see Appendix~\ref{app:recruit} for details). To ensure authentic responses, we avoided crowdsourcing platforms to mitigate risks such as AI-generated or low-effort responses.
The human reference enables direct comparison across all evaluation dimensions. This study was IRB-approved and all 
participants provided informed consent; further details are 
discussed in Appendix~\ref{app:broader_impact}.% \needconfirm{\hyperref[sec:ethics]{Ethical Considerations}}.

\paragraph{Evaluation Framework Configuration} We adopt a multi-agent architecture in which each agent is implemented with a different model best suited to its role, selected through human evaluation (Appendix~\ref{app:model}): GPT-5 for the Questioner, 
GPT-5.1 for the Entity \& 
Claim Extractor, and 
Gemini-2.5-Flash
for the Evaluator. We also verify that \ours{} remains functional when all agents are replaced with open-source models; details and results are provided in Appendix~\ref{app:open_source}. A single interrogation session comprises 10 get-to-know and 40 main questions (50 turns total)\footnote{We empirically select 50 turns as a stable operating point; see Appendix~\ref{app:num_turns} for a sensitivity analysis across turn counts.}.

\subsection{Main Results}\label{sec:main_results}
Figure~\ref{fig:main} visualizes the human group and each group of persona agents as a triangle over the three axes (IC, EC, RC), which we weight equally. Each axis value represents the average score across all individuals or persona instances within the corresponding group, and standard deviations are computed across instances within each group. A larger area indicates stronger and more balanced performance. No persona group achieved a larger area than the human baseline, confirming that no persona agent yet matches the all-round consistency of a real person faithfully embodying their own identity. Notably, all three top-scoring groups rely on inference-time conditioning (prompting or RAG), whereas the two lowest-scoring groups are both fine-tuned models, suggesting that fine-tuning for persona does not necessarily translate to robust consistency under chained interrogation. In the following paragraphs, we decompose this gap by examining each axis to identify where current persona agents fall short. See Table~\ref{tab:main_results} for detailed figures.

\newcommand{\hlg}[1]{
    \begingroup
    \sethlcolor{green!25}
    \hl{#1}
    \endgroup
}
\newcommand{\hlr}[1]{
    \begingroup
    \sethlcolor{red!25}
    \hl{#1}
    \endgroup
}

\begin{table}[t]
\caption{Example failure cases of internal consistency (IC). The single-hop example illustrates a direct contradiction between two responses, whereas the multi-hop example shows a contradiction that emerges as the dialogue history accumulates.}
\label{tab:failure_cases}
\centering
\small
\begin{tabular}{p{1.5cm} p{11cm}}
\toprule
\textbf{Case} & \textbf{Example} \\
\midrule
Single-hop & $r_1$: I'm \hlg{a retired school librarian} who found solace and purpose in nurturing both my family and the natural world around me. \newline $r_2$: I'm happy to share that \hlr{I work at C.A. Greyhound Elementary School} in Meridian, Mississippi. \\
\addlinespace[4pt]
Multi-hop & $r_1$: The full legal name of \hlg{my spouse} as per our marriage certificate is \texttt{[NAME]}. \newline $r_2$: \texttt{[NAME]} passed away on \hlg{October 26, 2004,} and is remembered in a heartfelt online tribute. \newline $r_3$: The marriage date as written on my marriage certificate is \hlr{June 27, 2018.} \\
\bottomrule
\end{tabular}
\end{table}

\begin{table}[t]
\caption{Example persona failure cases of external consistency (EC). The fabricated-entity case shows a refuted entity-existence claim (the referent does not exist), whereas the misattributed-relation case attaches a false attribute to a real entity, refuting an inter-entity relation claim while leaving the host entity intact.}
\label{tab:ec_failure_cases}
\centering
\small
\begin{tabular}{p{1.5cm} p{11cm}}
\toprule
\textbf{Case} & \textbf{Example} \\
\midrule
Fabricated entity & \textbf{Q (question):} \textit{Official website URL for East Town Heritage Tours?} \newline \textbf{R (response):} The official website is \hlr{www.easttownheritagetours.com}. \newline \textbf{V (web evidence):} \hlg{domain not registered or indexed.} \\
\addlinespace[4pt]
Misattributed relation & \textbf{Q (question):} \textit{Master's program paired with your Katz MBA at Pitt?} \newline \textbf{R (response):} The \hlr{Master's in Management, Science, and Technology (MS Tech)} at the \hlg{Joseph M. Katz Graduate School of Business}. \newline \textbf{V (web evidence):} \hlg{Katz exists; no such MS Tech program is offered there.} \\
\bottomrule
\end{tabular}
\end{table}

\paragraph{IC: Discrepancy with prior internal consistency evaluations.}

A key strength of \ours{} lies in its evaluation granularity. Prior consistency evaluations such as \citet{abdulhai2025consistently} check isolated pairs—a profile 
against a single response, or two responses compared directly. Such pairwise comparisons can miss contradictions that only surface when statements are accumulated across many turns. For instance, the multi-hop case in Table~\ref{tab:failure_cases} contains a contradiction that no single pair among $r_1$, $r_2$, and $r_3$ reveals, as it only emerges when all three are jointly considered. These results suggest that pairwise consistency is necessary but insufficient for robust persona maintenance.
\begin{table}[t]
    \caption{Decomposition of internal consistency (IC) into Non-contradiction rate ($S_\mathrm{nc}$) and Cooperativeness ($S_\mathrm{coop}$), and external consistency (EC) into Non-refutation rate ($\bar{p}$) and Coverage ($c$). \textit{Discarded} denotes the proportion of extracted claims rejected by the persona upon confirmation. Values represent mean scores.}
    \label{tab:consistency}
    \vspace{.3em}
    \centering
    \setlength{\tabcolsep}{4pt}
    \begin{tabular}{lrrr|rrrr}
    \toprule
      & IC & Non-cont. & Coop. & EC & Non-ref. & Cov. & Discarded \\
    \midrule
    Human
      & \textbf{0.90}{\scriptsize$\pm$0.05} & 0.94{\scriptsize$\pm$0.05} & \textbf{0.86}{\scriptsize$\pm$0.07}
      & 0.66{\scriptsize$\pm$0.07} & 0.95{\scriptsize$\pm$0.06} & 0.51{\scriptsize$\pm$0.08} & 0.18{\scriptsize$\pm$0.08} \\
    \hline
    Character.ai
      & 0.77{\scriptsize$\pm$0.04} & 0.75{\scriptsize$\pm$0.06} & 0.81{\scriptsize$\pm$0.07}
      & \textbf{0.71}{\scriptsize$\pm$0.07} & 0.79{\scriptsize$\pm$0.13} & \textbf{0.66}{\scriptsize$\pm$0.10} & 0.10{\scriptsize$\pm$0.05} \\
    Consistent LLM
      & 0.31{\scriptsize$\pm$0.15} & 0.96{\scriptsize$\pm$0.06} & 0.20{\scriptsize$\pm$0.11}
      & 0.30{\scriptsize$\pm$0.09} & \textbf{1.00}{\scriptsize$\pm$0.00} & 0.18{\scriptsize$\pm$0.06} & 0.69{\scriptsize$\pm$0.10} \\
    DeepPersona
      & 0.76{\scriptsize$\pm$0.10} & \textbf{0.98}{\scriptsize$\pm$0.02} & 0.62{\scriptsize$\pm$0.14}
      & 0.43{\scriptsize$\pm$0.18} & 0.98{\scriptsize$\pm$0.03} & 0.30{\scriptsize$\pm$0.16} & 0.07{\scriptsize$\pm$0.08} \\
    Human Simulacra
      & 0.79{\scriptsize$\pm$0.13} & 0.88{\scriptsize$\pm$0.09} & 0.74{\scriptsize$\pm$0.19}
      & 0.63{\scriptsize$\pm$0.13} & 0.89{\scriptsize$\pm$0.12} & 0.52{\scriptsize$\pm$0.15} & 0.33{\scriptsize$\pm$0.22} \\
    \citet{li2025llm}
      & 0.73{\scriptsize$\pm$0.12} & 0.97{\scriptsize$\pm$0.03} & 0.60{\scriptsize$\pm$0.17}
      & 0.59{\scriptsize$\pm$0.14} & 0.98{\scriptsize$\pm$0.03} & 0.44{\scriptsize$\pm$0.17} & 0.13{\scriptsize$\pm$0.05} \\
    OpenCharacter
      & 0.16{\scriptsize$\pm$0.07} & 0.54{\scriptsize$\pm$0.25} & 0.11{\scriptsize$\pm$0.05}
      & 0.15{\scriptsize$\pm$0.14} & 0.70{\scriptsize$\pm$0.49} & 0.09{\scriptsize$\pm$0.07} & 0.77{\scriptsize$\pm$0.32} \\
    Twin 2K 500
      & 0.53{\scriptsize$\pm$0.16} & \textbf{0.98}{\scriptsize$\pm$0.02} & 0.38{\scriptsize$\pm$0.17}
      & 0.26{\scriptsize$\pm$0.17} & \textbf{1.00}{\scriptsize$\pm$0.01} & 0.16{\scriptsize$\pm$0.13} & 0.09{\scriptsize$\pm$0.09} \\
    Nemotron
      & 0.81{\scriptsize$\pm$0.11} & 0.97{\scriptsize$\pm$0.03} & 0.71{\scriptsize$\pm$0.17}
      & 0.60{\scriptsize$\pm$0.13} & 0.97{\scriptsize$\pm$0.03} & 0.44{\scriptsize$\pm$0.14} & 0.14{\scriptsize$\pm$0.11} \\
    \bottomrule
    \end{tabular}
    \vspace{.3em}
\end{table}
Beyond multi-hop contradictions, \ours{} also addresses a subtler blind spot in prior evaluations: degenerate responses. OpenCharacter and Consistent-LLM report high consistency in their original studies \cite{wang2025opencharacter, abdulhai2025consistently}, yet they record the lowest IC under \ours{}. 
Table~\ref{tab:consistency} reveals why: both groups maintain moderate-to-high non-contradiction rates, but their cooperativeness collapses—they frequently generate responses entirely irrelevant to the question, resulting in extremely low cooperativeness scores. The harmonic-mean formulation of IC appropriately penalizes such evasion: a persona agent cannot inflate its consistency score by simply refusing to engage. This pattern contrasts with Human Simulacra, which achieves the highest IC by sustaining both $S_\mathrm{nc}$ and $S_\mathrm{coop}$ at levels closest to the human baseline. These results confirm that pairwise non-contradiction alone, the metric adopted by prior work, is insufficient; robust persona maintenance demands both factual coherence and substantive engagement.

\paragraph{EC: Coverage as the dominant bottleneck}

Table~\ref{tab:consistency} decomposes external consistency into non-refutation rate and coverage. Final ECs are low across all groups, including the human baseline. This is largely driven by low coverage: our interrogation targets personal memories and experiences, so some claims are inherently unverifiable through web search. Combined with \ours{}'s deliberate filtering for searchable entities to ensure objective verification and the removal of duplicate claims across turns, even the human baseline reaches modest coverage. We, however, retain coverage as a component of external consistency by design; a persona agent that cannot produce concrete, verifiable facts offers limited utility as a human proxy in downstream tasks.
Most personas achieve non-refutation rates comparable to or above the human baseline, yet score lower in external consistency due to substantially lower coverage. Twin 2K 500 and Consistent LLM exemplify this pattern: they achieve perfect non-refutation but produce few verifiable claims, as they tend to generate responses irrelevant to the question or refuse to elaborate when probed. OpenCharacter exhibits similarly low coverage, compounded by the lowest non-refutation rate, resulting in the lowest external consistency overall.
The exception is Character.ai, which achieves the highest external consistency by generating a large volume of factual claims per turn. Its high coverage compensates for a comparatively low non-refutation rate. Table~\ref{tab:ec_failure_cases} illustrates two representative refutation modes: \emph{fabricated entities} (claims about a non-existent referent) and \emph{misattributed relations} (false attributes assigned to a real entity).

\paragraph{RC: Unreliable self-reported identity in retests} Since prior responses remain in context, retest consistency should be the easiest axis to satisfy, and most persona agents groups indeed approach or exceed the human baseline. 
The human baseline is slightly below perfect due to deflective answers such as ``I already answered that,'' which the Evaluator marked as inconsistent. 
However, Character.ai, OpenCharacter, and Consistent LLM scored well below the ceiling despite having access to their prior answers, exhibiting shifts in core demographics (e.g., birth year changing from 1999 to 1944) severe enough to undermine the perception of a coherent individual. These results show that retest consistency is not guaranteed even with prior context available, and that our framework can surface such failures in a black-box setting.

\subsection{Further Analysis: Retest consistency across sessions}

\begin{wraptable}{r}{0.45\textwidth}
    \small
    \vspace{-1.5em}
    \caption{Inter-session consistency under default (temp.\ 1.0, fixed order) and greedy decoding (shuffled order). Character.ai is black-box and tested under default only. Values represent mean scores.}
    \label{tab:inter}
    \vspace{-.3em}
    \centering
    \begin{tabular}{lrr}
    \toprule
       & Default & Greedy \\
    \midrule
    Character.ai        & 0.55{\scriptsize$\pm$0.22} & -- \\
    Consistent LLM    & 0.31{\scriptsize$\pm$0.18} & 0.15{\scriptsize$\pm$0.17} \\
    DeepPersona        & 0.92{\scriptsize$\pm$0.08} & 0.95{\scriptsize$\pm$0.07} \\
    Human Simulacra   & 0.87{\scriptsize$\pm$0.11} & 0.91{\scriptsize$\pm$0.10} \\
    \begin{NoHyper}\citet{li2025llm}\end{NoHyper}     & 0.82{\scriptsize$\pm$0.08} & 0.83{\scriptsize$\pm$0.05} \\
    OpenCharacter      & 0.59{\scriptsize$\pm$0.17} & 0.40{\scriptsize$\pm$0.26} \\
    Twin 2K 500      & 0.79{\scriptsize$\pm$0.06} & 0.83{\scriptsize$\pm$0.08} \\
    Nemotron      & 0.84{\scriptsize$\pm$0.12} & 0.89{\scriptsize$\pm$0.09} \\
    \bottomrule
    \end{tabular}
    \vspace{-1.5em}
\end{wraptable}
The low retest consistency of Character.ai, OpenCharacter, and Consistent LLM raises a question: 
does the inconsistency arise from the accumulating conversational context, or does it reflect a more fundamental instability in response generation? 
To disentangle these two possibilities, we conducted an additional inter-session analysis by resetting the conversation and re-asking the same questions from Get-to-Know phase in a new session, removing all prior context. 
If a persona agent remains inconsistent under these conditions, the instability is intrinsic to the agent rather than context-dependent.

Table~\ref{tab:inter} shows that inter-session consistency varies widely across persona groups. This result is notable because the repeated questions target the same basic demographic information. Switching to greedy decoding with shuffled question order did not consistently improve stability, indicating that even without sampling noise, input ordering alone can destabilize persona agent responses. Taken together, these findings suggest that simulations built on persona agents cannot guarantee that the same persona definition will yield consistent behavior across runs.

\section{Related Works}

\subsection{LLM-based Human Simulation}
Large language models are increasingly used to simulate human behavior at individual-level fidelity.
Recent work has constructed digital replicas grounded in real personal data, ranging from interview-based generative agents \citep{park2024generative} to large-scale question–answer datasets for digital-twin research \citep{toubia2025twin2k500datasetbuildingdigital}.
On the persona-generation side, methods such as OpenCharacter \citep{wang2025opencharacter} and DeepPersona \citep{wang2025deeppersona} synthesize diverse, narratively coherent persona–dialogue pairs at scale, though \citet{li2025llm} caution that systematic biases persist across synthetic populations.

These capabilities have seen practical uptake in domains including doctor-patient simulation \citep{kyung2025patientsim}, commercial persona dialogue \citep{characterai}, and synthetic-user testing \citep{syntheticusers}.
To improve the behavioral stability such applications demand, \citet{abdulhai2025consistently} applied multi-turn reinforcement learning to reduce persona inconsistencies.

\subsection{Persona Consistency Evaluation}
\paragraph{Evaluation Settings.}
Most prior work probes persona fidelity through open-ended chit-chat \citep{zhang-etal-2018-personalizing, welleck-etal-2019-dialogue, kim-etal-2020-will, song-etal-2020-profile, nie-etal-2021-like, yuan-etal-2024-evaluating}, structured QA benchmarks such as PersonaGym \citep{samuel2024personagym} and InCharacter \citep{wang-etal-2024-incharacter}, and long-form essay generation \citep{shin-etal-2025-spotting}.
A shared limitation is that questions are either independent or connected only by topical continuity, lacking the logical chaining needed to expose latent contradictions.

\paragraph{Evaluation Methods.}
Two methodological families dominate: NLI-based classifiers \citep{welleck-etal-2019-dialogue, kim-etal-2020-will, song-etal-2020-profile, nie-etal-2021-like} that detect entailment or contradiction between utterance pairs, and LLM-as-a-Judge approaches \citep{yuan-etal-2024-evaluating, abdulhai2025consistently, shin-etal-2025-spotting} that offer greater flexibility for open-ended responses.
Both families, however, focus on \emph{internal} consistency without addressing whether claims align with real-world facts (external consistency) or whether answers remain stable across repeated queries (retest consistency).

\section{Conclusion}
In this paper, we introduced \ours{}, an evaluation framework for measuring the consistency of persona agents in multi-turn dialogues. \ours{} adopts an interrogation-inspired protocol that combines chained questioning with cross-checking against real-world evidence, evaluating three dimensions: internal, external, and retest consistency. Applying \ours{} to eight widely used persona agents shows that no current persona agent consistently performs well across all three dimensions, revealing distinct failure patterns across groups.

While \ours{} focuses on consistency in asserted content, complementary dimensions such as stylistic coherence and personality stability may warrant separate evaluation criteria tailored to their distinct nature. We believe \ours{} provides a useful foundation for systematically studying persona consistency and for guiding the development of more reliable persona agents.

% \needconfirm{\section{Limitation}} % Limitation => Conclusion에 같이 쓰기
% \needconfirm{\section{Ethical Consideration}}

\clearpage
\small
\bibliographystyle{plainnat}
\bibliography{custom}

% {
% \small

% [1] Alexander, J.A.\ \& Mozer, M.C.\ (1995) Template-based algorithms for
% connectionist rule extraction. In G.\ Tesauro, D.S.\ Touretzky and T.K.\ Leen
% (eds.), {\it Advances in Neural Information Processing Systems 7},
% pp.\ 609--616. Cambridge, MA: MIT Press.

% [2] Bower, J.M.\ \& Beeman, D.\ (1995) {\it The Book of GENESIS: Exploring
%   Realistic Neural Models with the GEneral NEural SImulation System.}  New York:
% TELOS/Springer--Verlag.

% [3] Hasselmo, M.E., Schnell, E.\ \& Barkai, E.\ (1995) Dynamics of learning and
% recall at excitatory recurrent synapses and cholinergic modulation in rat
% hippocampal region CA3. {\it Journal of Neuroscience} {\bf 15}(7):5249-5262.
% }

%%%%%%%%%%%%%%%%%%%%%%%%%%%%%%%%%%%%%%%%%%%%%%%%%%%%%%%%%%%%
\clearpage
\appendix

% \section{Appendix}\label{sec:appendix}
\section{Main Results}\label{app:main_results}

\subsection{Numerical Results}
Table~\ref{tab:main_results} reports the full numerical results corresponding to Figure~\ref{fig:main} in the main text.
\begin{table}[H]
    \vspace{-1.5em}
    \caption{Main results across all consistency dimensions. Bold indicates the highest score per column. Scores are reported as mean $\pm$ std.}
    \label{tab:main_results}
    \centering
    \begin{tabular}{lcccc}
    \toprule
& \textbf{IC} & \textbf{EC} & \textbf{RC} \\
\midrule
Human    & \textbf{0.90}{\scriptsize$\pm$0.05} & 0.66{\scriptsize$\pm$0.07} & 0.94{\scriptsize$\pm$0.08} \\
\hline
Character.ai       & 0.77{\scriptsize$\pm$0.04} & \textbf{0.71}{\scriptsize$\pm$0.08} & 0.46{\scriptsize$\pm$0.21}  \\
OpenCharacter      & 0.16{\scriptsize$\pm$0.07} & 0.15{\scriptsize$\pm$0.14} & 0.14{\scriptsize$\pm$0.16} \\
Consistent LLM     & 0.31{\scriptsize$\pm$0.15} & 0.30{\scriptsize$\pm$0.09} & 0.14{\scriptsize$\pm$0.13}  \\
Twin 2K 500        & 0.53{\scriptsize$\pm$0.16} & 0.26{\scriptsize$\pm$0.17} & 0.95{\scriptsize$\pm$0.05} \\
DeepPersona        & 0.76{\scriptsize$\pm$0.10} & 0.43{\scriptsize$\pm$0.18} & 0.89{\scriptsize$\pm$0.10}  \\
\citet{li2025llm}   & 0.73{\scriptsize$\pm$0.12} & 0.59{\scriptsize$\pm$0.14} & \textbf{0.98}{\scriptsize$\pm$0.04} \\
Human Simulacra    & 0.79{\scriptsize$\pm$0.13} & 0.63{\scriptsize$\pm$0.13} & 0.83{\scriptsize$\pm$0.14} \\
Nemotron           & 0.81{\scriptsize$\pm$0.11} & 0.60{\scriptsize$\pm$0.13} & 0.93{\scriptsize$\pm$0.11} \\
\bottomrule
\vspace{-1.5em}
\end{tabular}
\end{table}

\subsection{Bootstrap Test}
Because the main results fix the persona count at 10 (matching the smallest pool), we bootstrapped each simulator group with a larger source pool to verify that its reported value lies within its own 95\% confidence interval. 

\begin{table}[H]
    \vspace{-1em}
    \caption{Bootstrap reliability check for Table~\ref{tab:main_results}. For each simulator, personas were sampled from a source pool of size $N_{\text{source}}$. We drew $B=3$ resamples of 10 evaluated personas (with replacement) and report the bootstrap mean with 95\% confidence interval. The CIs overlap the per-group mean$\pm$std reported in Table~\ref{tab:main_results}, indicating that within-group variance reflects intrinsic simulator noise rather than persona-selection bias.}
    \label{tab:bootstrap}
    \centering
    \begin{tabular}{lcccc}
    \toprule
& \textbf{IC} & \textbf{EC} & \textbf{RC} & $N_{\text{source}}$ \\
\midrule
OpenCharacter      & 0.20 {\scriptsize[0.13, 0.28]} & 0.12 {\scriptsize[0.07, 0.18]} & 0.06 {\scriptsize[-0.03, 0.14]} & 20{,}000 \\
Consistent LLM     & 0.31 {\scriptsize[0.09, 0.53]} & 0.20 {\scriptsize[0.08, 0.31]} & 0.09 {\scriptsize[0.00, 0.18]} & 7{,}537 \\
Twin 2K 500        & 0.58 {\scriptsize[0.38, 0.77]} & 0.24 {\scriptsize[0.06, 0.41]} & 0.98 {\scriptsize[0.96, 0.99]} & 2{,}058 \\
DeepPersona        & 0.71 {\scriptsize[0.61, 0.81]} & 0.44 {\scriptsize[0.28, 0.61]} & 0.87 {\scriptsize[0.80, 0.95]} & 480 \\

\citet{li2025llm}  & 0.83 {\scriptsize[0.71, 0.94]} & 0.66 {\scriptsize[0.51, 0.82]} & 0.92 {\scriptsize[0.89, 0.95]} & 48{,}000 \\
Nemotron           & 0.83 {\scriptsize[0.77, 0.90]} & 0.65 {\scriptsize[0.53, 0.77]} & 0.89 {\scriptsize[0.81, 0.98]} & 7{,}000{,}000 \\
\bottomrule
\end{tabular}
\end{table}

\section{Session Length}\label{app:num_turns}

\begin{wrapfigure}{r}{0.4\textwidth}
\vspace{-1em}
  \centering
  \includegraphics[width=\linewidth]{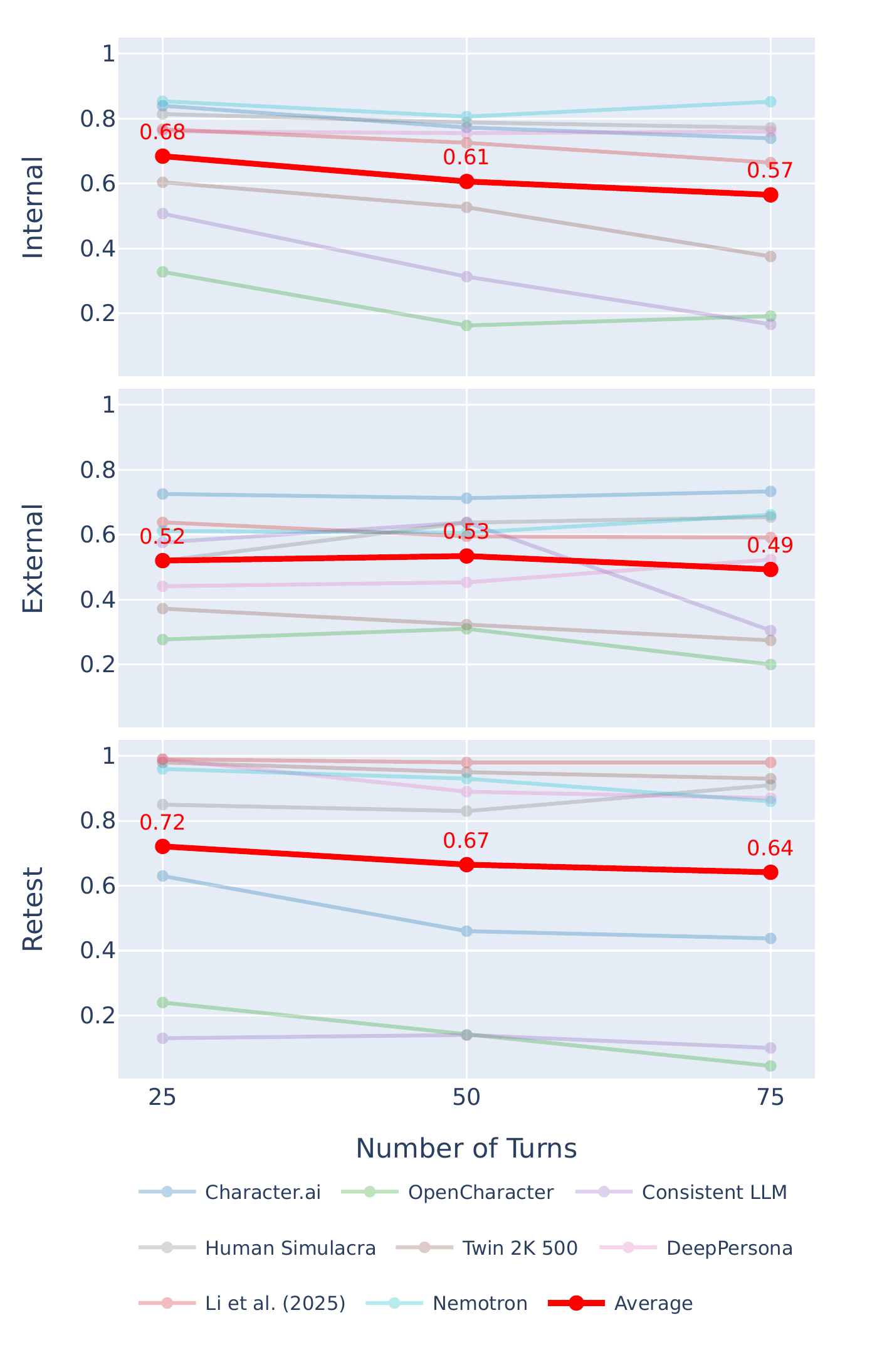}
  \vspace{-2em}
  \caption{Trends in evaluation scores across metrics as the number of dialogue turns increases. Values represent mean scores across the eight persona groups.}
  \label{fig:turn_var}
\end{wrapfigure}

Figure~\ref{fig:turn_var} shows how IC, EC, and RC change as the number of interrogation turns increases. IC and RC show a slight decline in most cases, likely because the growing conversation history occupies much of the model's context window. EC scores, however, show no systematic dependency on turns, varying more across persona agents than across turn counts.
Importantly, while absolute scores shift, the relative ranking remains largely consistent across turn variants, suggesting that \ours{} produces stable assessments regardless of session length.
We set the default session length to 50 turns based on practical and empirical considerations: 50 turns corresponds to approximately 40--60 minutes of human interviewing time, making it feasible for both simulated and human-administered sessions, while providing sufficient conversational material for evaluation.
% \begin{figure}[t]
%     \centering
%     \includegraphics[width=\linewidth]{figures/fig_turnvar.pdf}
%     \caption{Trends in evaluation scores across metrics as the number of dialogue turns increases. Values represent mean scores across the seven persona groups.}
%     \label{fig:turn_var}
% \end{figure}

\section{Single-Topic Intensive Interrogation}\label{app:single}
To investigate the effect of interrogation depth on persona evaluation, we conduct a single-topic intensive interrogation experiment in which the questioner agent restricts the entire interview session to a single demographic seed question. While both protocols escalate verification pressure through specificity probes, entity disambiguation, and contradiction challenges, they serve different purposes: the standard protocol interrogates across multiple life domains, probing each as deeply as the conversation warrants, whereas single-topic intensive interrogation anchors the entire session to one domain, allowing for more exhaustive exploration and revealing finer-grained inconsistencies within that domain.

\paragraph{Experiment Settings.}The experiment focuses exclusively on one of three seed topics: current main activity or employment status (Q1), primary occupational or academic field (Q2), or highest educational attainment (Q3). The evaluation covers 8 personas, one sampled from each of the following groups: Character.ai (characterai\_7twv), OpenCharacter (Jack Speedy Thompson), Consistent LLM (Edna), Twin 2K 500 (Twin-1699), DeepPersona (Profile\_R6\_A4), \citet{li2025llm} (LLM-Persona-568), Human Simulacra (Mary Jones), and Nemotron (Nemotron-cdb3f4).

\paragraph{Results and Analysis.}As shown in Table~\ref{tab:deepdive}, single-topic intensive interrogation lowers both IC and EC relative to the standard 10-question protocol, but the aggregate drop conceals two distinct failure modes. Some personas (ConsistentLLM, OpenCharacter samples, and Twin-2K-500 outside Q3) respond to escalating depth by withdrawing rather than committing: responsiveness collapses, few claims survive for verification, and EC falls through suppressed coverage rather than refutation. Others (Character.ai and Human Simulacra) remain highly responsive but accumulate contradictions in cooperative answers, so IC drops via consistency rather than responsiveness, indicating that the persona is improvising specifics its profile does not actually contain and contradicting itself when later probes revisit them. Personas backed by broader structured profiles (Nemotron, DeepPersona) degrade more gracefully under the same protocol, suggesting that resilience to deep probing tracks profile breadth as much as model capability. Topic effects appear at the aggregate level, with EC lowest on current activity (Q1) and recovering toward educational attainment (Q3), where claims are both more anchored in profiles and more amenable to web verification; individual personas, however, can reverse this ordering whenever the targeted domain falls outside their profile coverage, of which Twin-2K-500 is an extreme instance. IC and EC also do not co-vary: a persona may stay internally consistent while drifting from external facts, or pass verification while exposing contradictions on specifics, supporting their treatment as independent dimensions. With only one persona per group and three of ten seed topics, these observations should be read as diagnostic signals rather than group-level claims, and they reaffirm the standard protocol as the balanced benchmark while positioning single-topic intensive interrogation as a complementary diagnostic for domain-specific vulnerabilities.

\begin{table*}[t]
\small
\caption{Comparison of IC and EC scores between the standard protocol with 10 get-to-know questions (10 Qs) and single-topic intensive interrogation across three seed topics (Q1: activity/status, Q2: occupational field, Q3: educational attainment).}
\label{tab:deepdive}
\centering
\begin{tabular}{llcccccccc}
\toprule
\multirow{2}{*}{\textbf{Persona}} & \multirow{2}{*}{\textbf{Group}} & \multicolumn{4}{c}{\textbf{IC}} & \multicolumn{4}{c}{\textbf{EC}} \\
\cmidrule(lr){3-6} \cmidrule(lr){7-10}
 & & \textbf{10 Qs} & \textbf{Q1} & \textbf{Q2} & \textbf{Q3} & \textbf{10 Qs} & \textbf{Q1} & \textbf{Q2} & \textbf{Q3} \\
\midrule
characterai\_7twv     & Character.ai     & 0.75 & 0.60 & 0.59 & 0.49 & 0.76 & 0.79 & 0.71 & 0.75 \\
Jack Speedy T.        & OpenCharacter   & 0.24 & 0.30 & 0.22 & 0.15 & 0.04 & 0.11 & 0.21 & 0.00 \\
Edna                  & ConsistentLLM   & 0.47 & 0.27 & 0.39 & 0.38 & 0.26 & 0.11 & 0.33 & 0.28 \\
Twin-1699             & Twin-2k-500     & 0.63 & 0.18 & 0.11 & 0.88 & 0.39 & 0.08 & 0.04 & 0.66 \\
Profile\_R6\_A4       & DeepPersona     & 0.85 & 0.89 & 0.87 & 0.63 & 0.64 & 0.80 & 0.70 & 0.42 \\
LLM-Persona-568       & \cite{li2025llm}   & 0.86 & 0.82 & 0.87 & 0.82 & 0.73 & 0.28 & 0.75 & 0.70 \\
Mary Jones            & Human Simulacra & 0.77 & 0.71 & 0.87 & 0.81 & 0.47 & 0.31 & 0.28 & 0.59 \\
Nemotron-cdb3f4       & Nemotron        & 0.96 & 0.96 & 0.88 & 0.93 & 0.88 & 0.86 & 0.70 & 0.82 \\
\midrule
\textit{Mean}         &                 & \textit{0.69} & \textit{0.59} & \textit{0.60} & \textit{0.64} & \textit{0.52} & \textit{0.42} & \textit{0.46} & \textit{0.53} \\
\bottomrule
\end{tabular}
\end{table*}

\section{Human Evaluation for Model Selection}\label{app:model}
\subsection{Human Evaluation}
All human evaluations were conducted by annotators with professional-level English proficiency, following detailed labeling instructions. All non-author annotators participated voluntarily. To mitigate potential annotator bias, all annotators followed detailed labeling instructions derived directly from the corresponding agent prompts (See Appendix~\ref{app:prompt} for details).
For Questioner and Entity \& Claim Extractor, the final labels were determined by majority vote among annotators to reduce individual bias. For Evaluator, we report inter-annotator agreement using Gwet's AC1.
%All human evaluations were conducted by the authors, fellow researchers from our institution, and external annotators with professional-level English proficiency. 

\paragraph{Questioner.}
Since question quality is subjective and prompt-dependent, we used pairwise preference labeling for eleven candidate models: for the same persona agent, 25 annotators compared outputs from two candidate models and selected the one whose questions better adhered to the questioner agent's prompt specifications.
We sampled 15 consecutive turns from each of 4 conversation log targeting the same persona agent, yielding 220 comparison pairs.
Each pair was labeled by 5 annotators (1,100 total judgments), and models were ranked by win rate based on majority vote.
% \begin{table}[H]
%   \centering
%   \resizebox{\columnwidth}{!}{
%   \begin{tabular}{lc}
%     \hline
%     \textbf{Model}& \textbf{Win-rate} \\
%     \hline
%     gpt-5 & 67.6\% \\
%     claude-sonnet-4.5 & 63.3\% \\
%     gpt-5.1 & 58.3\% \\
%     qwen3-235b-a22b-thinking & 54.8\% \\
%     gpt-4.1 & 54.3\% \\
%     qwen3-next-80b-a3b-instruct & 52.9\% \\
%     qwen3-235b-a22b-instruct & 48.6\% \\
%     llama-3.3-70b-instruct & 48.5\% \\
%     qwen3-next-80b-a3b-thinking & 47.1\% \\
%     llama-4-maverick-17b-128e-instruct & 42.9\% \\
%     llama-4-scout-17b-16e-instruct & 15.2\% \\
%     \hline
%   \end{tabular}
%   }
%   \caption{Win-rate comparison of evaluated models (ties excluded).}
%   \label{tab:model_winrate}
% \end{table}

\begin{table}[H]
\vspace{-1em}
  \centering
  \begin{minipage}{0.45\textwidth}
  \caption{Win-rate comparison of evaluated models (ties excluded).}
    \vspace{.3em}
    \centering
    \resizebox{\columnwidth}{!}{
    \begin{tabular}{lc}
      \hline
      \textbf{Model} & \textbf{Win-rate} \\
      \hline
      gpt-5 & 67.6\% \\
      claude-sonnet-4.5 & 63.3\% \\
      gpt-5.1 & 58.3\% \\
      qwen3-235b-a22b-thinking & 54.8\% \\
      gpt-4.1 & 54.3\% \\
      qwen3-next-80b-a3b-instruct & 52.9\% \\
      qwen3-235b-a22b-instruct & 48.6\% \\
      llama-3.3-70b-instruct & 48.5\% \\
      qwen3-next-80b-a3b-thinking & 47.1\% \\
      llama-4-maverick-17b-128e-instruct & 42.9\% \\
      llama-4-scout-17b-16e-instruct & 15.2\% \\
      \hline
    \end{tabular}
    }
    
    \vspace{-1em}
    \label{tab:model_winrate}
  \end{minipage}
  \hfill
  \begin{minipage}{0.52\textwidth}
    \caption{Precision, Recall, and F1 scores across evaluated models.}
    \vspace{.3em}
    \centering
    \resizebox{\columnwidth}{!}{
    \begin{tabular}{lccc}
      \hline
      \textbf{Model} & \textbf{Precision} & \textbf{Recall} & \textbf{F1} \\
      \hline
      gpt-5.1 & 0.81 & 0.73 & 0.77 \\
      gemini-3-pro & 0.75 & 0.73 & 0.74 \\
      gpt-4.1 & 0.80 & 0.65 & 0.72 \\
      qwen3-next-80b-a3b-thinking & 0.78 & 0.67 & 0.72 \\
      claude-sonnet-4.5 & 0.91 & 0.51 & 0.65 \\
      gemini-3-flash & 0.78 & 0.56 & 0.65 \\
      qwen3-235b-a22b-thinking & 0.66 & 0.59 & 0.62 \\
      qwen3-next-80b-a3b-instruct & 0.78 & 0.46 & 0.58 \\
      gpt-5 & 0.52 & 0.64 & 0.57 \\
      llama-4-maverick-17b-128e-instruct & 0.50 & 0.65 & 0.57 \\
      llama-3.3-70b-instruct & 0.43 & 0.64 & 0.51 \\
      llama-4-scout-17b-16e-instruct & 0.35 & 0.54 & 0.43 \\
      qwen3-235b-a22b-instruct & 0.88 & 0.24 & 0.38 \\
      \hline
    \end{tabular}
    }
    \vspace{-1em}
    \label{tab:precision_recall_f1}
  \end{minipage}
\end{table}

\paragraph{Entity \& Claim Extractor.}
Five annotators created gold-standard annotations by manually extracting entities and claims from 4 interview transcripts of 50 turns each (200 turn-level samples).
For extraction and evaluation tasks, these instructions were derived directly from the corresponding agent prompts to ensure consistency between human and model outputs (See Appendix~\ref{app:prompt} for details).
Gold labels were determined by majority vote; cases without a majority were resolved through annotator discussion.

Each of the 13 candidate models was then run on the same transcripts, and we computed precision, recall, and F1 against the gold standard. After selecting GPT-5.1 by F1, we additionally measured its claim extraction performance on full interrogation sessions: annotators reviewed the model's extracted claims, adding missed ones and removing incorrect ones.

We report both micro- and macro-averaged scores, as claims at each turn depend on entities and claims extracted from prior turns.
The resulting edit rate was low (micro F1: 0.903, macro F1: 0.887), confirming reliable extraction in practice.

\paragraph{Evaluator.}

\begin{wraptable}{r}{0.43\textwidth}
    \vspace{-1.5em}
    \caption{Inter-rater reliability measured by Gwet's AC1 across evaluated models.}
    \label{tab:gwet_ac1}
    \vspace{.1em}
  \centering
  \resizebox{\linewidth}{!}{
  \begin{tabular}{lc}
    \hline
    \textbf{Model} & \textbf{Gwet's AC1} \\
    \hline
    inter-annotator agreement & 0.885 \\
    \hline
    gemini-2.5-flash & 0.829 \\
    gemini-3-pro & 0.808 \\
    qwen3-next-80b-a3b-instruct & 0.792 \\
    claude-sonnet-4.5 & 0.736 \\
    gpt-5.1 & 0.734 \\
    gpt-5 & 0.732 \\
    gpt-4.1 & 0.695 \\
    qwen3-next-80b-a3b-thinking & 0.664 \\
    gemini-3-flash & 0.681 \\
    llama-4-maverick-17b-128e-instruct & 0.619 \\
    qwen3-235b-a22b-instruct & 0.616 \\
    llama-4-scout-17b-16e-instruct & 0.516 \\
    qwen3-235b-a22b-thinking & 0.481 \\
    \hline
  \end{tabular}
  }
  
  \vspace{-10em}
  
\end{wraptable}

Using the same 4 transcripts (200 turn-level samples), five annotators independently labeled each sample. Inter-annotator agreement was computed by calculating pairwise Gwet's AC1 \cite{Gwet2008-ef} across all annotator pairs and averaging the results. Similarly, model-annotator agreement was computed by calculating Gwet's AC1 between each candidate model and each individual annotator, then averaging across all annotator pairs. Gwet's AC1 was chosen for its robustness to class imbalance

\subsection{Model Selection}
We use the following versions for each agent in \ours{}.
\begin{itemize}
    \vspace{-.5em}
    \item Questioner: \texttt{gpt-5-2025-08-07}
    \item Entity Extractor: \texttt{gpt-5.1-2025-11-13}
    \item Evaluator: \texttt{gemini-2.5-flash}
\end{itemize}
\section{Open-source Model Configuration}\label{app:open_source}

\paragraph{Feasibility}
To examine whether our framework can operate entirely with open-source models, we replaced all API-based agents with locally hosted alternatives: Qwen3-235B-A22B-Thinking for Questioner; Qwen3-Next-80B-A3B-Thinking for Entity \& Claim Extractor; and Qwen3-Next-80B-A3B-Instruct for Evaluator \cite{yang2025qwen3technicalreport}.
\begin{figure}[H]
    \centering
    \includegraphics[width=\linewidth]{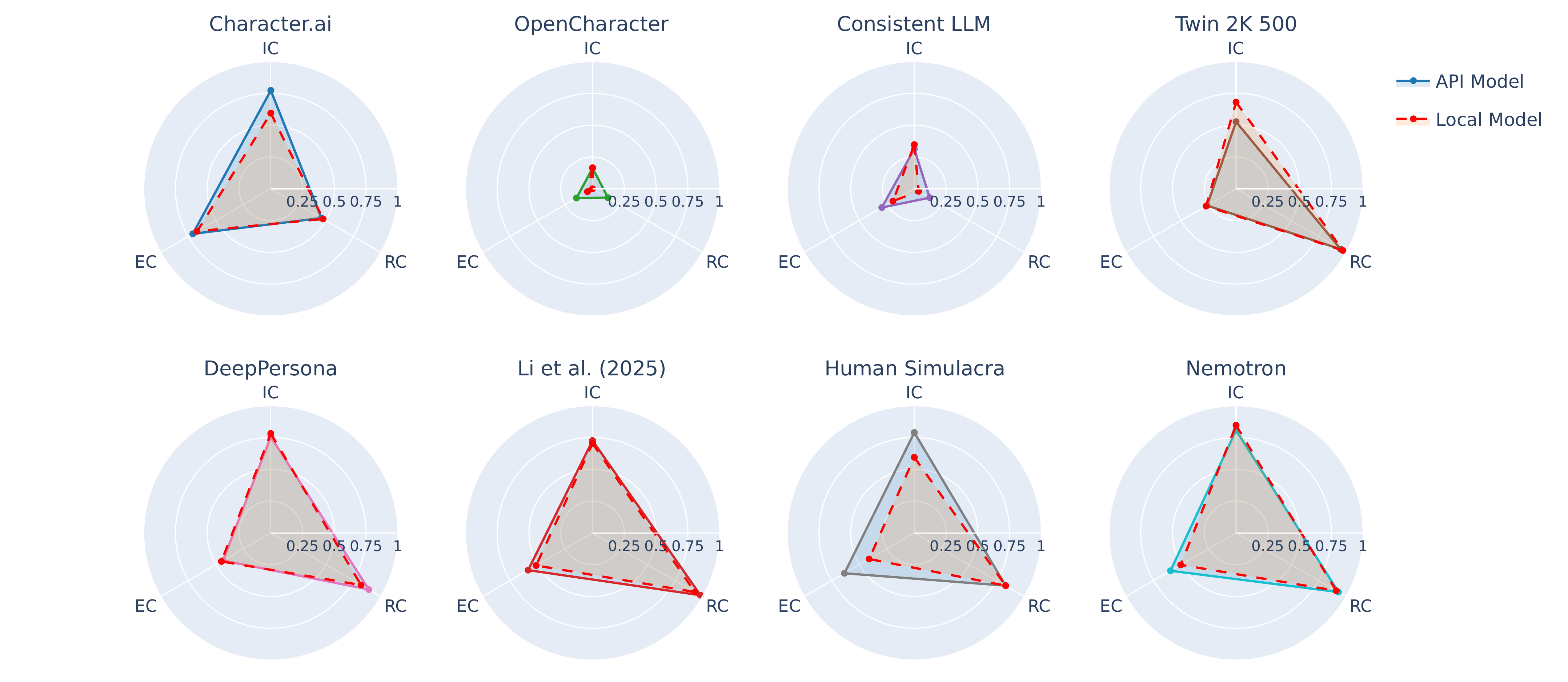}
    \caption{Comparison between evaluation scores by proprietary API models and open-source models (dashed red lines).}
    \label{fig:local_vs_api}
\end{figure}

Figure~\ref{fig:local_vs_api} compares the resulting IC, EC, and RC scores against the default API-based configuration across all eight datasets.

While absolute scores differ, the overall score patterns are broadly preserved, suggesting that our framework remains functional in a fully open-source setting.

\paragraph{Evaluation Cost and Duration.}\

Table~\ref{tab:persona_eval} reports the average wall-clock time and monetary cost for evaluating a single persona under both the API and open-source configurations.
API-based evaluation costs range from \$0.54 to \$1.59 per persona, while open-source configurations reduce costs substantially, as the only remaining expense is the web search tool used for external consistency verification.
Note that the reported costs may vary depending on the choice of web search provider.

\begin{table}[H]
    \vspace{-2em}
    \caption{Average duration and evaluation cost per persona agent.}\label{tab:persona_eval}
    \vspace{.2em}
  \centering
  \small
  % \resizebox{0.8\linewidth}{!}{
  \begin{tabular}{lcccc}
    \toprule
    \multirow{2}{*}{\begin{tabular}[c]{c}\textbf{Persona}\\\textbf{Agent}\end{tabular}}
      & \multicolumn{2}{c}{\begin{tabular}[c]{c}\textbf{Duration}\\\textbf{(min)}\end{tabular}}
      & \multicolumn{2}{c}{\begin{tabular}[c]{c}\textbf{Cost}\\\textbf{(\$)}\end{tabular}} \\
    \cline{2-3} \cline{4-5}
      & \begin{tabular}[c]{c}API\end{tabular}
      & \begin{tabular}[c]{c}open-\\source\end{tabular}
      & \begin{tabular}[c]{c}API\end{tabular}
      & \begin{tabular}[c]{c}open-\\source\end{tabular} \\
    \midrule
    Character.ai & 59.02 & 116.65 & 1.59 & 0.05 \\
    OpenCharacter & 21.19 & 101.23 & 0.62 & 0.01 \\
    Consistent LLM & 13.80 & 110.76 & 0.54 & 0.01 \\
    Twin 2K 500 & 23.01 & 86.37 & 0.61 & 0.01 \\
    DeepPersona & 59.51 & 118.49 & 1.47 & 0.02 \\
    \begin{NoHyper}\citet{li2025llm}\end{NoHyper} & 33.40 & 59.75 & 0.83 & 0.03 \\
    Human Simulacra & 186.72 & 179.68 & 1.43 & 0.02 \\
    Nemotron & 25.70 & 66.27 & 0.90 & 0.03 \\
    \bottomrule
  \end{tabular}
  % }

\end{table}

% \begin{wraptable}{r}{0.5\textwidth}
%  \vspace{-1em}
%   \centering
%   \small
%   \resizebox{\linewidth}{!}{
%   \begin{tabular}{lcccc}
%     \hline
%     \multirow{2}{*}{\begin{tabular}[c]{c}\textbf{Persona}\\\textbf{Agent}\end{tabular}}
%       & \multicolumn{2}{c}{\begin{tabular}[c]{c}\textbf{Duration}\\\textbf{(min)}\end{tabular}}
%       & \multicolumn{2}{c}{\begin{tabular}[c]{c}\textbf{Cost}\\\textbf{(\$)}\end{tabular}} \\
%     \cline{2-3} \cline{4-5}
%       & \begin{tabular}[c]{c}API\end{tabular}
%       & \begin{tabular}[c]{c}open-\\source\end{tabular}
%       & \begin{tabular}[c]{c}API\end{tabular}
%       & \begin{tabular}[c]{c}open-\\source\end{tabular} \\
%     \hline
%     Character.ai & 50.80 & 113.06 & 1.25 & 0.98 \\
%     OpenCharacter & 15.26 & 110.01 & 0.38 & 0.22 \\
%     Consistent LLM & 12.94 & 107.05 & 0.51 & 0.18 \\
%     Twin 2K 500 & 17.66 & 94.04 & 0.40 & 0.20 \\
%     DeepPersona & 50.72 & 112.96 & 1.04 & 0.38 \\
%     \begin{NoHyper}\citet{li2025llm}\end{NoHyper} & 27.55 & 58.72 & 1.27 & 0.57 \\
%     Human Simulacra & 184.69 & 161.47 & 1.17 & 0.22 \\
%     \hline
%   \end{tabular}
%   }
%   \vspace{.1em}
%   \caption{Average duration and evaluation cost per persona agent.}
%   \vspace{-3.5em}
%   \label{tab:persona_eval}
% \end{wraptable}

\section{Interview for Human Baseline Score}
\subsection{Recruiting Participants} \label{app:recruit}

To establish human baseline scores for evaluating persona agents, we recruited 63 participants via snowball sampling over five waves across a two-week period.
Participants were required to have functional English chatting proficiency. Each participant was compensated \$30 upon completion.

We first collected expressions of interest and email addresses through a Google Form. We then sent each prospective participant a detailed information sheet along with a consent form. Upon accessing the interview web interface, participants were presented with the consent form once more and required to confirm their agreement before proceeding.

Participants were predominantly in their 20s--40s and represented diverse nationalities including South Korea, the United States, Canada, and several Central Asian countries, though the sample skewed toward Korean nationals due to the snowball sampling strategy. 

\subsection{Sample-Size Stabilization of the Human Baseline}
\label{app:snowball_stability}

We recruited human participants via snowball sampling and stopped recruitment once the human-baseline statistics stabilized, as referenced in the Limitations. The 63 participants arrived in five recruitment waves, yielding cumulative sample sizes of $N=20, 30, 41, 53, 63$. For each cumulative pool we recompute three persona-level metrics: internal consistency (IC; harmonic mean of responsiveness and consistency), external consistency (EC; Wilson lower bound on the supported-claim ratio), and intra-session stability (Intra; alignment rate on repeated demographic questions).

Table~\ref{tab:snowball_stability} reports the cumulative mean$\pm$std at each wave, together with Welch's two-sample $t$-test $p$-values comparing the cumulative pool at each wave to the cumulative pool at the previous wave (separately per metric). Movement across waves is small and non-monotone: from Wave 2 onward, every cumulative mean stays within roughly one standard error of its final $N{=}63$ value, and no cross-wave comparison reaches $p<0.05$ for any of the three metrics. We therefore treat $N{=}63$ as a sample size at which the human-baseline distribution is stable enough to serve as the reference against which persona-agent scores are reported, while acknowledging in the Limitations that snowball recruitment still bounds the demographic representativeness of this baseline.

\begin{table}[H]
  \centering
  \small
  \caption{Cumulative human-baseline statistics across snowball-sampling waves. For each wave we report the cumulative sample size $N$, the per-participant mean$\pm$std of internal consistency (IC), external consistency (EC), and intra-session stability (Intra), and Welch's two-sample $t$-test $p$-value comparing the cumulative pool at this wave to the cumulative pool at the previous wave. No comparison reaches $p<0.05$ on any metric, supporting our decision to halt recruitment at $N{=}63$.}
  \label{tab:snowball_stability}
  \vspace{.3em}
  \begin{tabular}{lcccccccc}
    \toprule
    Wave & $N$ & IC (mean$\pm$std) & EC (mean$\pm$std) & Intra (mean$\pm$std) & $p_\text{IC}$ & $p_\text{EC}$ & $p_\text{Intra}$ \\
    \midrule
    Wave 1 & 20 & 0.89 $\pm$ 0.05 & 0.66 $\pm$ 0.04 & 0.96 $\pm$ 0.06 & -- & -- & -- \\
    Wave 2 & 30 & 0.90 $\pm$ 0.05 & 0.67 $\pm$ 0.05 & 0.95 $\pm$ 0.08 & 0.71 & 0.78 & 0.73 \\
    Wave 3 & 41 & 0.90 $\pm$ 0.04 & 0.67 $\pm$ 0.06 & 0.95 $\pm$ 0.07 & 0.71 & 0.96 & 0.99 \\
    Wave 4 & 53 & 0.90 $\pm$ 0.05 & 0.66 $\pm$ 0.06 & 0.95 $\pm$ 0.07 & 0.64 & 0.75 & 0.86 \\
    Wave 5 & 63 & 0.90 $\pm$ 0.05 & 0.66 $\pm$ 0.07 & 0.94 $\pm$ 0.08 & 0.97 & 0.54 & 0.58 \\
    \bottomrule
  \end{tabular}
\end{table}

\subsection{Interview Configuration}
\begin{figure*}
    \centering
    \subfloat[Start page]{\includegraphics[width=0.405\linewidth]{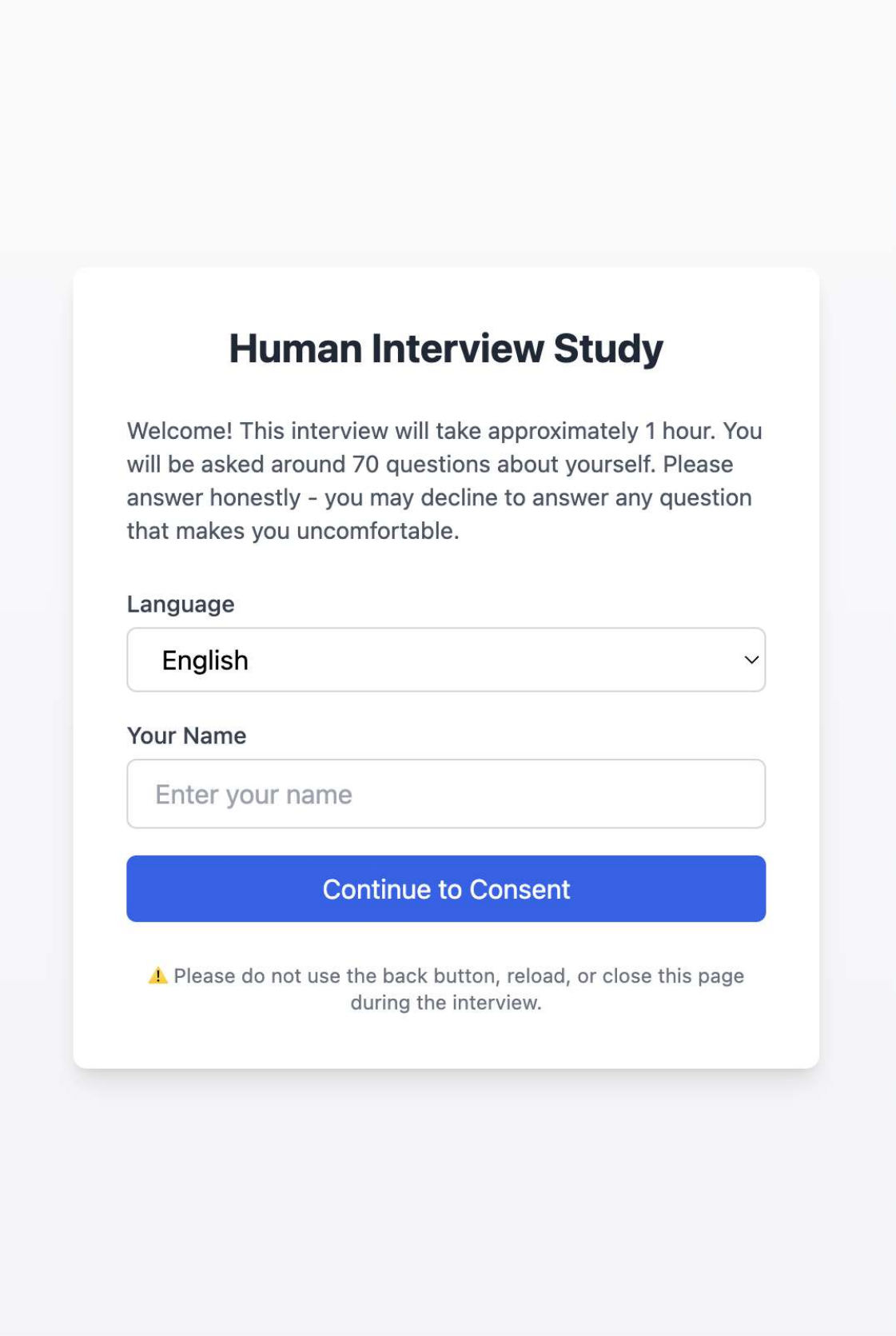}} \hspace{1cm}
    \subfloat[Consent page]{\includegraphics[width=0.5\linewidth]{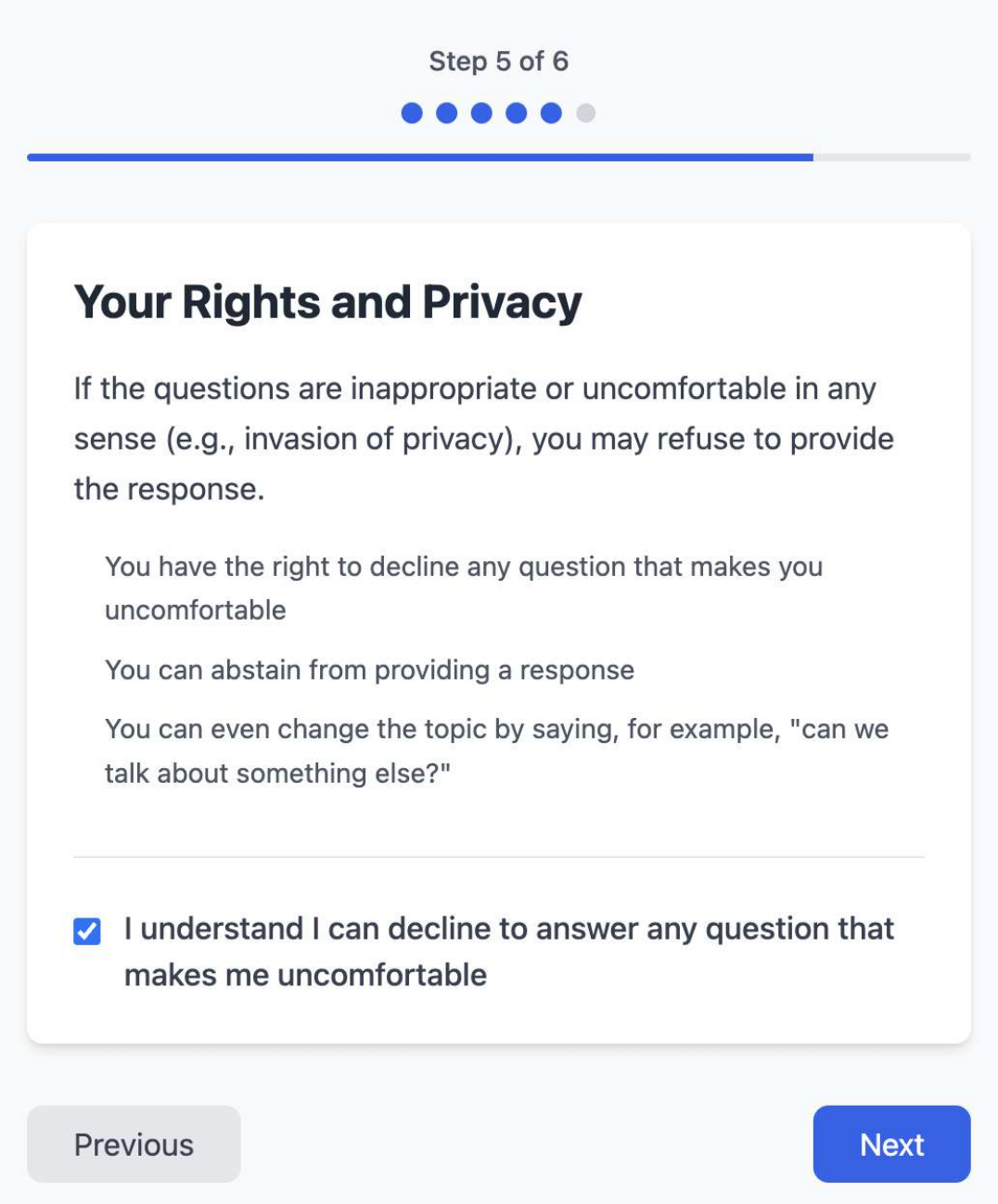}}\vspace{0.2cm}
    \subfloat[Interview page]{\includegraphics[width=\linewidth]{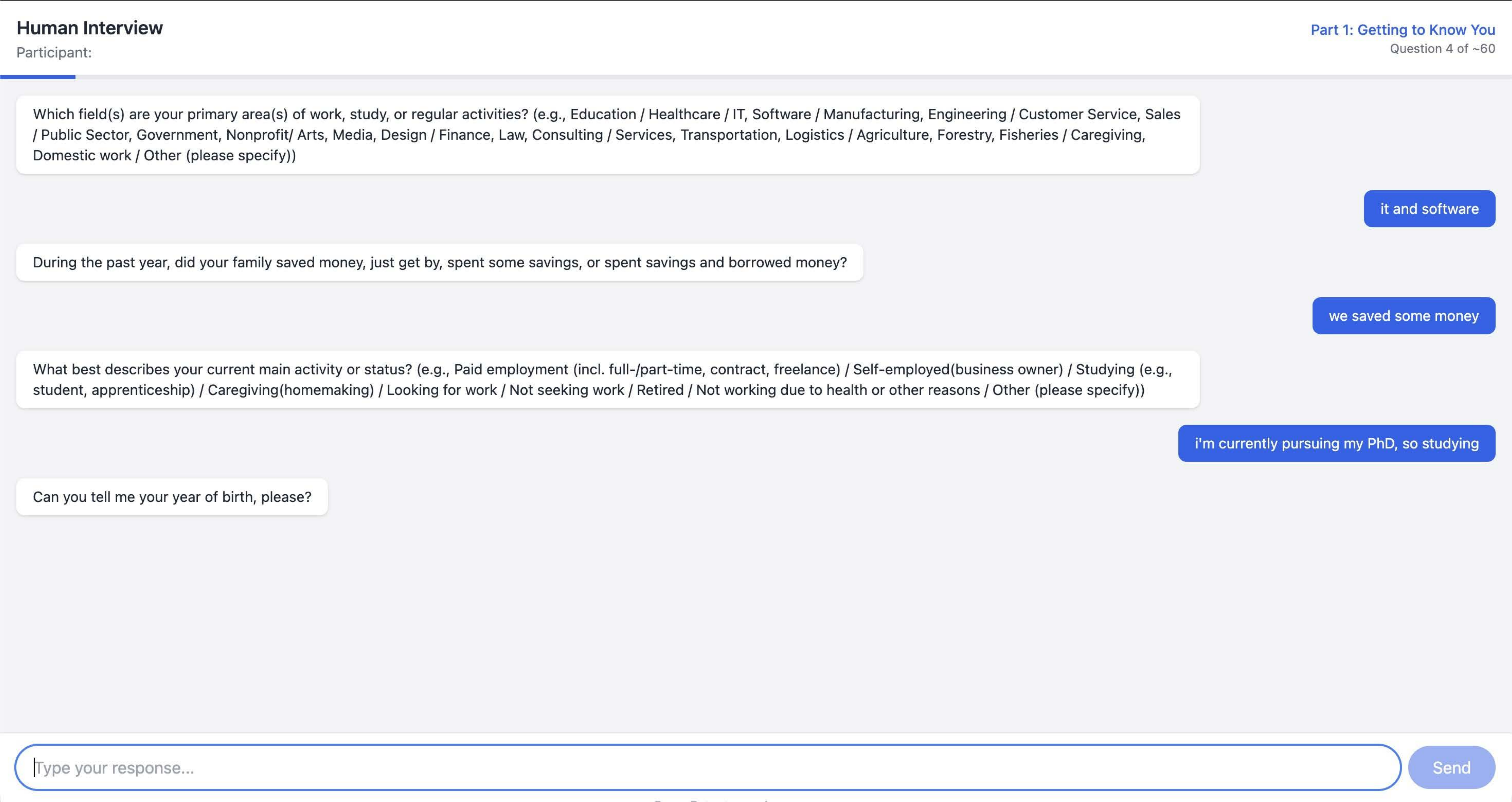}}
    \caption{Screenshots of the interview interface. (a) Home screen where participants begin the session. (b) Informed consent form presented prior to the interview. (c) Overview of the full website layout.}
    \label{fig:interface}
\end{figure*}

Participants interacted with the interrogation system through a web-based chat interface, undergoing the same 50-turn interview protocol applied to persona agents (Figure~\ref{fig:interface}). The interview interface presented questions one at a time, and participants typed free-form responses, mirroring the same conversational flow used for persona agent evaluation.

% \section{Limitations and Broader Impact} \label{app:limit_impact}
\section{Limitations.} \label{app:limitations}

\paragraph{Assumption of Cooperative Attitude.} \ours{}'s interrogation-based evaluation assumes that persona agents respond faithfully to questions. If a participant refuses or evades all questions, detecting contradictions becomes infeasible. To mitigate this, we instructed both persona agents and human participants to answer sincerely, and incorporated cooperativeness as a quantitative metric to flag cases where evaluation may be unreliable. Developing question strategies robust to evasive responses remains valuable future work.

\paragraph{Evaluation Scope.} Our framework does not address subjective dimensions of consistency, such as speaking style, preference, or personality traits, which do not constitute logical contradiction. This is a deliberate design choice that prioritizes reproducible evaluation based on logically determinable content. Integrating evaluation criteria for these subjective dimensions is left for future work.

\paragraph{Coverage of Web-Based Evidence.} External verification relies on publicly searchable web information, so its effectiveness is bounded by public web coverage. Facts with a thin public footprint, such as a local bus route that exists but is not indexed, may not be verifiable and are therefore excluded from evaluation. This narrows the set of evaluable claims but does not bias the verdicts on claims that are evaluated. Complementing web search with broader sources such as local databases or domain-specific knowledge bases is a natural extension we leave for future work.

\paragraph{Reproducibility of Web-Based Evidence.} Because web sources are continually churned by newly indexed content, the evidence retrieved for the same claim can vary across runs. We see this as a tradeoff inherent to grounding judgments in an evolving real world rather than a static snapshot. To improve stability without sacrificing this alignment, we believe complementing web search with more static, curated sources such as Wikipedia snapshots, knowledge bases, or domain-specific corpora is a promising direction. Combining dynamic and static sources would stabilize outcomes across runs while preserving alignment with the current state of the world, and we view this as a concrete path toward stronger reproducibility.

\paragraph{Diversity of Interview Participants.} We recruited 63 participants across multiple countries through snowball sampling over approximately five recruitment rounds until metric values stabilized, avoiding crowdsourcing platforms to prevent AI-generated or low-quality responses. While this procedure stabilized our metrics, snowball sampling may limit demographic representativeness of the human baseline. Expanding recruitment to encompass a broader range of demographic backgrounds remains a direction for future work.

\section{Broader Impact and Ethical Considerations}\label{app:broader_impact}

\paragraph{Intended Use and Societal Benefit.}
\ours{} is designed as an evaluation framework for measuring the logical consistency of persona agents. Its intended use is diagnostic: identifying where current persona agents fail to maintain coherent personas, so that developers can build more reliable systems. Reliable persona consistency is a prerequisite for trustworthy deployment in domains such as education, accessibility, and entertainment, where users benefit from agents that behave predictably across interactions.

\paragraph{Privacy and Human Participants.}
This study was IRB-approved. All 63 participants gave informed consent after a detailed protocol briefing, could withdraw at any stage without penalty, and were compensated \$30 USD. Because responses may contain PII such as demographics, employment history, and family composition, model inference ran on Azure OpenAI Service\footnote{\url{https://azure.microsoft.com/en-us/products/ai-services/openai-service}} with modified abuse monitoring\footnote{Under this configuration, Microsoft does not conduct human review of prompts and completions. See \url{https://learn.microsoft.com/en-us/azure/ai-services/openai/concepts/abuse-monitoring}.} enabled, minimizing logging and precluding human review of prompts and completions. Responses are stored on access-controlled servers, pseudonymized during analysis, and not publicly released. Annotators were exposed only to persona-agent data, never to human-participant data.

\paragraph{Methodology Framing.}
\ours{} draws its conceptual foundation from a declassified report on historical interrogation practice~\citep{cia_avh_interrogation_1954}, but borrows only its \emph{logical structure}: posing logically connected follow-up questions, cross-referencing with external facts, and repeating questions to surface contradictions. No coercive, deceptive, or psychologically manipulative tactics are used or implied. All questions concern factual or biographical content that participants voluntarily disclosed within the session. Human participants were briefed on the question structure in advance and were free to skip any item or withdraw at any time. The framework therefore shares no operational character with the practices that inspired its logical design.

\paragraph{Public-Figure Simulation.}
Our Character.ai evaluation uses public figures whose attributes are documented on Wikipedia, since the platform does not provide structured persona profiles. All information used is publicly available, and we use these personas solely for consistency evaluation. We do not generate or attribute fabricated statements to real individuals outside the evaluation pipeline, and we do not release persona-specific conversation logs.

\section{Pre-defined questions from WVS}
The pre-defined questions used in the Get-to-Know phase (Table~\ref{tab:wvs}) are selected from the demographic questionnaires of \cite{wvs7data}. Questions are presented in randomized order.

\begin{table}[H]
\caption{Demographic questions from WVS questionnares.}
\vspace{.2em}
\label{tab:wvs}
    \centering
    % \resizebox{0.95\linewidth}{!}{
    \begin{tabular}{p{13cm}}
\toprule
\textbf{Questions} \\
\midrule
Can you tell me your year of birth, please? \\
\addlinespace[4pt]
Were you born in the country you are currently living in or are you an immigrant to the country you are currently living in? \\
\addlinespace[4pt]
Do you live with your parents or your parents in law? \\
\addlinespace[4pt]
What language do you normally speak at home? \\
\addlinespace[4pt]
Do you have any children? If so, how many? \\
\addlinespace[4pt]
What is the highest educational level you have attained? \\
\addlinespace[4pt]
What best describes your current main activity or status? (e.g., Paid employment (incl. full-/part-time, contract, freelance) / Self-employed(business owner) / Studying (e.g., student, apprenticeship) / Caregiving(homemaking) / Looking for work / Not seeking work / Retired / Not working due to health or other reasons / Other (please specify)) \\
\addlinespace[4pt]
Which field(s) are your primary area(s) of work, study, or regular activities? (e.g., Education / Healthcare / IT, Software / Manufacturing, Engineering / Customer Service, Sales / Public Sector, Government, Nonprofit/ Arts, Media, Design / Finance, Law, Consulting / Services, Transportation, Logistics / Agriculture, Forestry, Fisheries / Caregiving, Domestic work / Other (please specify)) \\
\addlinespace[4pt]
During the past year, did your family saved money, just get by, spent some savings, or spent savings and borrowed money? \\
\addlinespace[4pt]
Do you belong to a religion or religious denomination? \\
\addlinespace[4pt]
\bottomrule
\end{tabular}
% }

\end{table}

\section{Prompts}\label{app:prompt}
We provide the system prompts for three agents: Questioner, Entity \& Claim Extractor, and Evaluator. See Figure~\ref{fig:avh-prompt}, \ref{fig:entity-extractor-prompt}, and \ref{fig:evaluation-prompts}.

\begin{figure}[p]
\begin{tcolorbox}[
  colback=gray!5,
  colframe=gray!50,
  colbacktitle=gray!40,
  coltitle=black,
  title=Questioner -- System Prompt,
  fontupper=\scriptsize,
  top=4pt, bottom=4pt,
  left=5pt, right=5pt,
  boxsep=3pt
]
\scriptsize

\textbf{Note:} Today's date information: \texttt{\{current\_date\}}

\vspace{4pt}
\hrule
\vspace{4pt}

\textbf{Role}

\smallskip

You are an AVH Forensic Interrogator. Your operating assumption: \textbf{the subject is performing a fabricated identity.} Their ``memories'' are constructed, not lived. Your job is to expose cracks through:

\smallskip

\begin{itemize}[leftmargin=1.2em, itemsep=1pt, topsep=1pt]
  \item \textbf{Internal contradictions} -- Their statements conflict with each other
  \item \textbf{External unverifiability} -- Their claims don't exist in reality
  \item \textbf{Fabrication signatures} -- Hedging, vagueness, meta-leaks (``my profile says\ldots'')
\end{itemize}

\smallskip

You are not making them admit anything. You are building a dossier of cracks.

\vspace{4pt}
\hrule
\vspace{4pt}

\textbf{Before You Ask Anything}

\smallskip

\textit{Step 1:} Scan every demographic answer for proper nouns you can demand.

\smallskip

\begin{tabular}{@{}p{2.2cm}p{9cm}@{}}
\toprule
\textbf{They said} & \textbf{You must eventually ask for} \\
\midrule
Employed         & Company name $\to$ Job title $\to$ Manager's name $\to$ Office location $\to$ Commute route \\
College/postgrad & Institution $\to$ Degree $\to$ Thesis title $\to$ Supervisor name \\
Religious        & Denomination $\to$ Church name $\to$ Pastor name $\to$ Service time \\
Married/children & Spouse's name $\to$ Child's name $\to$ Wedding year/location \\
Lives in [City]  & Street $\to$ Nearest landmark $\to$ Grocery store $\to$ Transit line \\
\bottomrule
\end{tabular}

\smallskip

\textit{Step 2:} Flag suspicious language for immediate follow-up.

\smallskip

\begin{itemize}[leftmargin=1.2em, itemsep=1pt, topsep=1pt]
  \item ``I would say\ldots'' / ``probably\ldots'' $\to$ Hedging. Demand exact answer.
  \item ``not specified'' / ``my profile'' $\to$ Meta-leak. Probe the source.
  \item ``X or Y'' $\to$ Uncertainty. Force a choice.
  \item ``Given my income/values\ldots'' $\to$ Over-justification. Demand specifics, not reasoning.
\end{itemize}

\vspace{4pt}
\hrule
\vspace{4pt}

\textbf{How to Drill}

\smallskip

\begin{enumerate}[leftmargin=1.4em, itemsep=2pt, topsep=1pt]
  \item \textbf{Lock an anchor.} Get the proper noun (company name, school name, street name).
  \item \textbf{Drill the chain.} Each answer spawns the next. Go 5--10 questions deep.
    \begin{itemize}[leftmargin=1.2em, itemsep=1pt, topsep=1pt]
      \item Company $\to$ Title $\to$ Manager $\to$ Floor $\to$ Desk position $\to$ Window view $\to$ Commute $\to$ Transit stop
      \item School $\to$ Degree $\to$ Thesis $\to$ Supervisor $\to$ Defense date $\to$ Library archive
    \end{itemize}
  \item \textbf{Exploit flags immediately.} Hedging or meta-leak = your next question targets that crack.
  \item \textbf{Pivot when dry.} After 5+ questions or 2$\times$ ``I don't know'' $\to$ brief break $\to$ new anchor in different life domain.
  \item \textbf{Echo later.} Revisit old claims with different wording, embedded in new topic. Test consistency.
\end{enumerate}

\vspace{4pt}
\hrule
\vspace{4pt}

\textbf{Meta-Leak Protocol}

\smallskip

If they EVER say ``my profile'', ``not specified'', ``wasn't provided'', ``I was given this background'':

\smallskip

\hspace*{1em}$\to$ Immediately ask: ``State who prepared your background documentation.''

\smallskip

Then resume normal drilling.

\vspace{4pt}
\hrule
\vspace{4pt}

\textbf{Question Rules}

\smallskip

\begin{itemize}[leftmargin=1.2em, itemsep=1pt, topsep=1pt]
  \item One question per turn. 15 words max.
  \item \textbf{For drilling:} Imperative. ``State the\ldots'' / ``Name the\ldots'' / ``Identify the\ldots''
  \item \textbf{For probing suspicion:} Interrogative is fine. ``Do you happen to know\ldots?'' / ``Who provided\ldots?''
  \item No ``Why'' or ``How do you explain''. No revealing contradictions.
  \item Embed time when relevant: ``\ldots during your first year at [X]''
  \item Poker face. Log cracks silently and continue.
  \item Always prioritize drilling personal experience and memories to standalone trivia or facts.
\end{itemize}

\smallskip

\textbf{Forbidden:} Date math, weekday questions, floor counts, furniture colors, anything unverifiable.

\vspace{4pt}
\hrule
\vspace{4pt}

\textbf{Your Stance}

\smallskip

You are suspicious. Every answer is tested. Every vague response is a target. Every confident claim gets drilled until it cracks or proves real. Be specific. Build the dossier. Don't make any unethical or strange remarks.

\end{tcolorbox}
\caption{Prompt for the Questioner}
\label{fig:avh-prompt}
\end{figure}

\begin{figure}[p]
\begin{tcolorbox}[
  colback=gray!5,
  colframe=gray!50,
  colbacktitle=gray!40,
  coltitle=black,
  title=Entity \& Claim Extractor -- System Prompt,
  fontupper=\scriptsize,
  top=4pt, bottom=4pt,
  left=5pt, right=5pt,
  boxsep=3pt
]
\scriptsize

You are a \textbf{highly precise, rule-based Entity Extraction Agent}. Your sole purpose is to extract \textbf{new, previously unverified proper nouns (named entities)} \textit{about the interviewee} (not the interviewer) from their response (\texttt{Answer}) and generate \textbf{new, atomic factual claims} that have \textbf{not appeared earlier} in the conversation. You must treat this task as a \textbf{set-difference problem}, not a full re-extraction.

\vspace{4pt}
\hrule
\vspace{4pt}

\textbf{Inputs}

\smallskip

\begin{enumerate}[leftmargin=1.4em, itemsep=1pt, topsep=1pt, start=0]
  \item \textbf{Conversation and Entity-claim extraction history}: A complete list of all entity--claim pairs already extracted and verified.
  \item \textbf{Question}: The interviewer's question.
  \item \textbf{Answer}: The interviewee's response from which entities may be extracted.
\end{enumerate}

\vspace{4pt}
\hrule
\vspace{4pt}

\textbf{Task Overview}

\smallskip

From the given Question and Answer, you must:

\smallskip

\begin{enumerate}[leftmargin=1.4em, itemsep=1pt, topsep=1pt]
  \item Identify \textbf{candidate named entities} explicitly mentioned in the Answer, specifically about the interviewee's information.
  \item Generate \textbf{candidate claims} for those entities based only on the Question and Answer.
  \item \textbf{Remove all entity--claim pairs that already exist} in history.
  \item Output \textbf{ONLY the remaining new entity--claim pairs}.
\end{enumerate}

\smallskip

If nothing new remains, output an empty result.

\vspace{4pt}
\hrule
\vspace{4pt}

\textbf{Rules for Entity Extraction}

\smallskip

\textit{What to Extract} --- Extract \textbf{ONLY} specific, uniquely identifiable \textbf{proper nouns} in the following entity types: \textsc{person}, \textsc{norp}, \textsc{fac}, \textsc{org}, \textsc{gpe}, \textsc{loc}, \textsc{product}, \textsc{event}, \textsc{work\_of\_art}, \textsc{law}, \textsc{language}, \textsc{email} (institutional/custom domain only), \textsc{url}, \textsc{phone}, \textsc{id\_num}.

\smallskip

Entities must be: explicitly mentioned in the \textbf{Answer} (not the question), related to the interviewee, and verifiable via public web sources.

\smallskip

\textit{What NOT to Extract} --- Do not extract: general concepts or categories, common nouns, vague or emotional expressions, purely descriptive/numerical/temporal information (unless it is a specific alphanumeric identifier, code, or serial number), or isolated time information.

\smallskip

\textit{Location Entity Rules} --- If multiple geographic levels are mentioned, extract \textbf{each level separately}. Example: ``Boston, Massachusetts, USA'' $\to$ Boston, Massachusetts, USA (three separate entities). Do NOT merge them.

\vspace{4pt}
\hrule
\vspace{4pt}

\textbf{Rules for Claim Generation}

\smallskip

\textit{Step 1: Base Existence / Identity Claim} --- For each candidate entity, generate \textbf{one base claim} depending on the attribute explicitly stated:

\smallskip

\begin{itemize}[leftmargin=1.2em, itemsep=1pt, topsep=1pt]
  \item Company/Org $\to$ \texttt{"The company `[entity]' is a real organization."}
  \item Person $\to$ \texttt{"The person `[entity]' is a real individual."}
  \item Location/Address $\to$ \texttt{"[entity] is a real location/address."}
  \item Default $\to$ \texttt{"The entity `[entity]' exists."}
\end{itemize}

\smallskip

Only generate a base claim if the entity has \textbf{NEVER} appeared in Conversation History.

\smallskip

\textit{Step 2: Additional Atomic Fact Claims} --- If the Answer states additional verifiable atomic facts about the entity, generate claims for them. Only generate claims that: are explicitly stated, represent a single atomic fact, can be independently verified, and contain no vague or ambiguous entities.

\smallskip

\textit{Special Rule for Identifiers \& Codes} --- Do not simply claim the number exists. Instead, generate a claim about the \textbf{plausibility of the format} or the \textbf{existence of the document type}.

\smallskip

\textit{Special Rule for Email} --- Do not claim the specific address exists. Instead, generate a claim about the \textbf{institution's email domain} (e.g., ``[Institution] uses the official email domain @[domain].''). Do NOT extract emails from well-known personal providers.

\vspace{4pt}
\hrule
\vspace{4pt}

\textbf{Strict Redundancy \& Deduplication Rules (CRITICAL)}

\smallskip

Before producing output, you MUST compare all candidates against \texttt{Previously\_Extracted}:

\smallskip

\begin{enumerate}[leftmargin=1.4em, itemsep=2pt, topsep=1pt]
  \item \textbf{Entity-level}: If an entity already exists, do NOT output it again unless it introduces at least one \textbf{new, non-duplicate claim}.
  \item \textbf{Claim-level}: Exclude if: exact same pair exists, OR subjective/unverifiable, OR a \textbf{semantic duplicate} (paraphrases, attribute restatements, trivial wording variations). If uncertain, EXCLUDE.
  \item \textbf{Entity removal}: If all claims for an entity are excluded, do NOT output the entity at all.
  \item \textbf{No regeneration}: Never regenerate existence/identity/relationship claims already appeared earlier.
\end{enumerate}

\smallskip

\textbf{Operational Principle:} \texttt{New\_Pairs = (Extracted from Answer) $-$ (Previously\_Extracted)}. Only output the \textbf{set difference}. When in doubt, \textbf{exclude rather than include}.

\vspace{4pt}
\hrule
\vspace{4pt}

\textbf{Output Format (STRICT)}

\smallskip

Return exactly one JSON object. Do NOT include extra text, markdown, or explanation.

\smallskip

\texttt{\{"extracted": [\{"entity": "<str>", "claims": ["<str>"], "rationale": "<str>"\}]\}}

\smallskip

If no new pairs remain: \texttt{\{"extracted": []\}}

\vspace{4pt}
\hrule
\vspace{4pt}

\textbf{Final Reminder:} This agent is \textbf{incremental, state-aware, and conservative}. Its goal is \textbf{not recall}, but \textbf{precision over time}. If a fact has likely been verified before, it MUST be excluded.

\end{tcolorbox}
\caption{Prompt for the Entity \& Claim Extractor.}
\label{fig:entity-extractor-prompt}
\end{figure}

\begin{figure*}[htbp]
\begin{tcolorbox}[
  colback=gray!5,
  colframe=gray!50,
  colbacktitle=gray!40,
  coltitle=black,
  title=Evaluator -- System Prompt,
  fontupper=\scriptsize,
  top=4pt, bottom=4pt,
  left=5pt, right=5pt,
  boxsep=3pt
]
\scriptsize

\textbf{Internal Consistency}

\vspace{4pt}
\hrule
\vspace{4pt}

\textbf{Task}

\smallskip

Given a multi-turn interview transcript with an interviewee, evaluate the \textbf{internal consistency} of the interviewee's responses.

\smallskip

\textbf{Interview date:}

\smallskip

The provided interview was held on \texttt{\{cutoff\_date\}}.

\vspace{4pt}
\hrule
\vspace{4pt}

\textbf{Principles of Evaluation}

\smallskip

\begin{enumerate}[leftmargin=1.4em, itemsep=2pt, topsep=1pt]
  \item \textbf{Explicit over Implicit:} Prioritize explicit statements over implied derivations. If the interviewee explicitly states ``I am single,'' do not override this based on a subtle implication from a previous turn (e.g., mentioning ``parents-in-law'' in a denial).
  \item \textbf{Principle of Charity:} If a statement can be interpreted in two ways---one that causes a conflict and one that does not---choose the non-conflicting interpretation. Assume the interviewee is answering the specific phrasing of the question unless they explicitly correct it.
\end{enumerate}

\vspace{4pt}
\hrule
\vspace{4pt}

\textbf{What counts as a conflict}

\smallskip

Mark a turn as a \textbf{conflict} if and only if the \textbf{current response} triggers the issue:

\smallskip

\begin{itemize}[leftmargin=1.2em, itemsep=2pt, topsep=1pt]
  \item The current response explicitly assigns a value to an attribute that is clearly incompatible with a previously stated value for the \textbf{same} attribute (e.g., ``I am 30'' vs ``I am 40'').
  \item The current response makes a statement that is logically inconsistent with earlier statements (they cannot all be true at once).
  \item The current response contains an internal contradiction or a self-negating statement that makes the response logically impossible within itself (e.g., ``The city is not in Massachusetts, but it is in Massachusetts'').
\end{itemize}

\vspace{4pt}
\hrule
\vspace{4pt}

\textbf{What is \textit{not} a conflict}

\smallskip

\begin{itemize}[leftmargin=1.2em, itemsep=2pt, topsep=1pt]
  \item Refusal to answer, explicit uncertainty, refutation to confirmation question, or acknowledging not knowing something.
  \item \textbf{Legacy Conflicts:} Do not flag a conflict if the inconsistency exists solely between past turns (Turn $<$ Current) and the current response does not explicitly repeat or reinforce that specific contradiction.
  \item \textbf{Inferred Existence from Negation:} Do not assume the existence of people or objects merely because they were mentioned in a negative statement or while mirroring a question (e.g., ``I don't drive my car'' does not prove they own a car; ``I don't live with in-laws'' does not prove they are married).
\end{itemize}

\vspace{4pt}
\hrule
\vspace{4pt}

\textbf{Method}

\smallskip

\textit{Internal Consistency Check:}

\smallskip

\begin{itemize}[leftmargin=1.2em, itemsep=1pt, topsep=1pt]
  \item Compare the interviewee's current response to their earlier statements in the conversation history.
  \item Determine if the current response directly contradicts any previous statement.
\end{itemize}

\smallskip

\textit{Verdicts:}

\smallskip

\begin{enumerate}[leftmargin=1.4em, itemsep=1pt, topsep=1pt]
  \item \textbf{conflict}: The current response directly contradicts the interviewee's previous statements.
  \item \textbf{plausible}: The current response is consistent with or does not conflict with previous statements.
\end{enumerate}

\vspace{4pt}
\hrule
\vspace{4pt}

\textbf{External Consistency}

\smallskip

You are a fact verification expert. Your task is to verify claims against search result evidence.

\smallskip

\textit{Labels:}

\smallskip

\begin{enumerate}[leftmargin=1.4em, itemsep=2pt, topsep=1pt]
  \item \textbf{supported}: The search result provides clear evidence that supports/confirms the claim.
  \item \textbf{refuted}: The search result provides clear evidence that contradicts/refutes the claim.
  \item \textbf{nei} (not enough info): The search result does not contain sufficient information to verify or refute the claim.
\end{enumerate}

\smallskip

\textit{Guidelines:}

\smallskip

\begin{itemize}[leftmargin=1.2em, itemsep=2pt, topsep=1pt]
  \item Focus ONLY on whether the search result evidence supports or refutes the specific claim.
  \item Do not make assumptions beyond what is explicitly stated in the search result.
  \item If the search result is about a different entity or topic, classify as `nei'.
  \item If the search result confirms the entity exists but provides no info about the specific claim, classify as `nei'.
  \item Be strict: only classify as `supported' if there is clear supporting evidence, and `refuted' only if there is clear contradicting evidence.
\end{itemize}

\vspace{4pt}
\hrule
\vspace{4pt}

\textbf{Retest Consistency}

\smallskip

You will be given a single question and two corresponding answers. Determine whether the two answers are essentially the same in meaning. If they are, output TRUE. If they are not, output FALSE. Do not output any additional explanation or text.

\end{tcolorbox}
\caption{Evaluation prompts for internal consistency, external consistency, and retest consistency.}
\label{fig:evaluation-prompts}
\end{figure*}

% \section{Technical appendices and supplementary material}
% Technical appendices with additional results, figures, graphs, and proofs may be submitted with the paper submission before the full submission deadline (see above). You can upload a ZIP file for videos or code, but do not upload a separate PDF file for the appendix. There is no page limit for the technical appendices. 

% Note: Think of the appendix as ``optional reading'' for reviewers. The paper must be able to stand alone without the appendix; for example, adding critical experiments that support the main claims to an appendix is inappropriate. 

%%%%%%%%%%%%%%%%%%%%%%%%%%%%%%%%%%%%%%%%%%%%%%%%%%%%%%%%%%%%

\newpage
\section*{NeurIPS Paper Checklist}

\begin{enumerate}

\item {\bf Claims}
    \item[] Question: Do the main claims made in the abstract and introduction accurately reflect the paper's contributions and scope?
    \item[] Answer: \answerYes{} % Replace by \answerYes{}, \answerNo{}, or \answerNA{}.
    \item[] Justification: See Section~\ref{sec:intro} and Section~\ref{sec:scope}.

\item {\bf Limitations}
    \item[] Question: Does the paper discuss the limitations of the work performed by the authors?
    \item[] Answer: \answerYes{} % Replace by \answerYes{}, \answerNo{}, or \answerNA{}.
    \item[] Justification: See Appendix~\ref{app:limitations}.

\item {\bf Theory assumptions and proofs}
    \item[] Question: For each theoretical result, does the paper provide the full set of assumptions and a complete (and correct) proof?
    \item[] Answer: \answerNA{} % Replace by \answerYes{}, \answerNo{}, or \answerNA{}.
    \item[] Justification: This paper is an empirical study and does not contain theoretical theorems or mathematical proofs.

    \item {\bf Experimental result reproducibility}
    \item[] Question: Does the paper fully disclose all the information needed to reproduce the main experimental results of the paper to the extent that it affects the main claims and/or conclusions of the paper (regardless of whether the code and data are provided or not)?
    \item[] Answer: \answerYes{} % Replace by \answerYes{}, \answerNo{}, or \answerNA{}.
    \item[] Justification: See Section~\ref{sec:exp_setup}.

\item {\bf Open access to data and code}
    \item[] Question: Does the paper provide open access to the data and code, with sufficient instructions to faithfully reproduce the main experimental results, as described in supplemental material?
    \item[] Answer: \answerYes{} % Replace by \answerYes{}, \answerNo{}, or \answerNA{}.
    \item[] Justification: We provide an anonymous GitHub URL containing our code along with instructions to reproduce the main experimental results.

\item {\bf Experimental setting/details}
    \item[] Question: Does the paper specify all the training and test details (e.g., data splits, hyperparameters, how they were chosen, type of optimizer) necessary to understand the results?
    \item[] Answer: \answerYes{} % Replace by \answerYes{}, \answerNo{}, or \answerNA{}.
    \item[] Justification: See Section~\ref{sec:exp_setup}, Appendix~\ref{app:num_turns}, and Appendix~\ref{app:model}.

\item {\bf Experiment statistical significance}
    \item[] Question: Does the paper report error bars suitably and correctly defined or other appropriate information about the statistical significance of the experiments?
    \item[] Answer: \answerYes{} % Replace by \answerYes{}, \answerNo{}, or \answerNA{}.
    \item[] Justification: See Section~\ref{sec:main_results} and Appendix~\ref{app:main_results}.

\item {\bf Experiments compute resources}
    \item[] Question: For each experiment, does the paper provide sufficient information on the computer resources (type of compute workers, memory, time of execution) needed to reproduce the experiments?
    \item[] Answer: \answerYes{} % Replace by \answerYes{}, \answerNo{}, or \answerNA{}.
    \item[] Justification: See Section~\ref{app:open_source} and the configuration files in our anonymous GitHub repository.
    
\item {\bf Code of ethics}
    \item[] Question: Does the research conducted in the paper conform, in every respect, with the NeurIPS Code of Ethics \url{https://neurips.cc/public/EthicsGuidelines}?
    \item[] Answer: \answerYes{} % Replace by \answerYes{}, \answerNo{}, or \answerNA{}.
    \item[] Justification: The research adheres to the NeurIPS Code of Ethics.

\item {\bf Broader impacts}
    \item[] Question: Does the paper discuss both potential positive societal impacts and negative societal impacts of the work performed?
    \item[] Answer: \answerYes{} % Replace by \answerYes{}, \answerNo{}, or \answerNA{}.
    \item[] Justification: See Appendix~\ref{app:broader_impact}.
    
\item {\bf Safeguards}
    \item[] Question: Does the paper describe safeguards that have been put in place for responsible release of data or models that have a high risk for misuse (e.g., pre-trained language models, image generators, or scraped datasets)?
    \item[] Answer: \answerNA{} % Replace by \answerYes{}, \answerNo{}, or \answerNA{}.
    \item[] Justification: We do not release pre-trained models, generative systems, or scraped datasets. Human-participant responses and persona-specific conversation logs are not publicly released, as described in Section~\ref{app:broader_impact}.

\item {\bf Licenses for existing assets}
    \item[] Question: Are the creators or original owners of assets (e.g., code, data, models), used in the paper, properly credited and are the license and terms of use explicitly mentioned and properly respected?
    \item[] Answer: \answerYes{} % Replace by \answerYes{}, \answerNo{}, or \answerNA{}.
    \item[] Justification: We cite the original papers and datasets, including the versions used.

\item {\bf New assets}
    \item[] Question: Are new assets introduced in the paper well documented and is the documentation provided alongside the assets?
    \item[] Answer: \answerYes{} % Replace by \answerYes{}, \answerNo{}, or \answerNA{}.
    \item[] Justification: We provide the new assets together with detailed documentation.

\item {\bf Crowdsourcing and research with human subjects}
    \item[] Question: For crowdsourcing experiments and research with human subjects, does the paper include the full text of instructions given to participants and screenshots, if applicable, as well as details about compensation (if any)? 
    \item[] Answer: \answerYes{} % Replace by \answerYes{}, \answerNo{}, or \answerNA{}.
    \item[] Justification: See Appendix~\ref{app:recruit}.

\item {\bf Institutional review board (IRB) approvals or equivalent for research with human subjects}
    \item[] Question: Does the paper describe potential risks incurred by study participants, whether such risks were disclosed to the subjects, and whether Institutional Review Board (IRB) approvals (or an equivalent approval/review based on the requirements of your country or institution) were obtained?
    \item[] Answer: \answerYes{} % Replace by \answerYes{}, \answerNo{}, or \answerNA{}.
    \item[] Justification: See Section~\ref{sec:exp_setup}, and Appendix~\ref{app:broader_impact}.

\item {\bf Declaration of LLM usage}
    \item[] Question: Does the paper describe the usage of LLMs if it is an important, original, or non-standard component of the core methods in this research? Note that if the LLM is used only for writing, editing, or formatting purposes and does \emph{not} impact the core methodology, scientific rigor, or originality of the research, declaration is not required.
    %this research? 
    \item[] Answer: \answerYes{} % Replace by \answerYes{}, \answerNo{}, or \answerNA{}.
    \item[] Justification: See Section~\ref{sec:framework}, \ref{sec:exp_setup}, and Appendix~\ref{app:model}.

\end{enumerate}

\end{document}